\documentclass[letterpaper, 10 pt, journal, twoside]{IEEEtran}

\usepackage{fixmath}
\usepackage{mathtools,amssymb}
\usepackage[dvipsnames]{xcolor}
\usepackage{color}
\usepackage{graphicx}
\usepackage{mathptmx}
\usepackage{epstopdf}
\usepackage{float}
\usepackage[T1]{fontenc} 
\usepackage[utf8]{inputenc} 
\usepackage{lmodern}
\usepackage[english]{babel} 

\usepackage{tikz}
\usepackage{mathptmx} 
\usepackage{times} 
\usepackage[ruled,vlined]{algorithm2e}
\usepackage{siunitx}
\usepackage{calc}
\usepackage{t1enc}
\usepackage{latexsym}
\usepackage{keyval}
\usepackage{ifthen}
\usepackage{moreverb}
\usepackage[noshell,siunitx]{gnuplottex}
\usepackage{xparse}
\usepackage{subcaption}

\usepackage[noabbrev]{cleveref}
\usepackage{diagbox}

\usepackage{multirow}
\usepackage{hhline}
\usepackage{tikzscale}
\usepackage{breqn}

\usepackage{amsmath,amsfonts} 
\newcommand{\subparagraph}{}
\usepackage{titlesec}
\titlespacing*{\section}
{0pt}{3mm}{1mm}
\titlespacing*{\subsection}
{0pt}{2mm}{1mm}
\usepackage{balance}
\ExplSyntaxOn
\DeclareExpandableDocumentCommand{\convertlen}{ O{cm} m }
 {
  \dim_to_decimal_in_unit:nn { #2 } { 1 #1 } cm
 }
\ExplSyntaxOff
\usepackage{xprintlen}
\IEEEoverridecommandlockouts

\author{Joshua Smith$^{1}$ and Michael Mistry$^{1}$%
\thanks{Manuscript received: September, 10\textsuperscript{th}, 2019; Revised December, 16\textsuperscript{th}, 2019;
Accepted January, 14\textsuperscript{th}, 2019.}
\thanks{This paper was recommended for publication by Editor Paolo Rocco upon evaluation of the Associate Editor and Reviewers' comments. This work has been supported by the following grants: EPSRC UK RAI Hub ORCA (EP/R026173/1) and NCNR (EPR02572X/1), CogIMon project in the EU Horizon 2020 (ICT-23-2014), THING project in the EU Horizon 2020 (ICT-2017-1), and by grant EP/L016834/1 for the University of Edinburgh RAS CDT from EPSRC.}
\thanks{$^{1}$Authors are with the School of Informatics,
University of Edinburgh, UK
{\tt\small Joshua.Smith@ed.ac.uk}, {\tt\small mmistry@inf.ed.ac.uk}}%

\thanks{Digital Object Identifier (DOI): see top of this page.}
}

\usepackage[absolute,showboxes]{textpos}

\TPMargin{5pt}

\newcommand{\copyrightstatement}{
    \begin{textblock}{0.84}(0.08,0.95)    
         \noindent
         \footnotesize
         \copyright 2020 IEEE.  Personal use of this material is permitted.  Permission from IEEE must be obtained for all other uses, in any current or future media, including reprinting/republishing this material for advertising or promotional purposes, creating new collective works, for resale or redistribution to servers or lists, or reuse of any copyrighted component of this work in other works.
    \end{textblock}
}

\makeatletter
\newcommand*\fsize{\dimexpr\f@size pt\relax}
\newcommand*{\getlength}[1]{\strip@pt#1}
\makeatother
\newlength\textRatio
\newlength\textRatioInv
\newlength\imagefontSize
\usetikzlibrary{positioning}
\usetikzlibrary{arrows} 
\usetikzlibrary{calc} 
\tikzstyle{block} = [draw,rectangle,thick,minimum height=2em,minimum width=2em]
\tikzstyle{sum} = [draw,circle,inner sep=0mm, outer sep=0pt,minimum size=2mm]
\tikzstyle{connector} = [->,thick]
\tikzstyle{line} = [thick]
\tikzstyle{branch} = [circle,inner sep=0pt, outer sep=0pt,minimum size=1mm,fill=black,draw=black]
\tikzstyle{guide} = [inner sep=0pt, outer sep=0pt]

\title{Online Simultaneous Semi-Parametric Dynamics Model Learning}

\begin{document}
\copyrightstatement
\setlength{\abovedisplayskip}{4pt}
\setlength{\belowdisplayskip}{4pt}
\setlength\abovecaptionskip{\baselineskip}
\setlength\belowcaptionskip{-0.7\baselineskip}
 
\def\@IEEEfigurecaptionsepspace{\vskip\abovecaptionskip\relax}%
\def\@IEEEtablecaptionsepspace{\vskip\abovecaptionskip\relax}%
\maketitle


\begin{abstract}
Accurate models of robots' dynamics are critical for control, stability, motion optimization, and interaction. Semi-Parametric approaches to dynamics learning combine physics-based Parametric models with unstructured Non-Parametric regression with the hope to achieve both accuracy and generalizability. In this paper, we highlight the non-stationary problem created when attempting to adapt both Parametric and Non-Parametric components simultaneously. We present a consistency transform designed to compensate for this non-stationary effect, such that the contributions of both models can adapt simultaneously without adversely affecting the performance of the platform. Thus, we are able to apply the Semi-Parametric learning approach for continuous iterative online adaptation, without relying on batch or offline updates. We validate the transform via a perfect virtual model as well as by applying the overall system on a Kuka LWR IV manipulator. We demonstrate improved tracking performance during online learning and show a clear transference of contribution between the two components with a learning bias towards the Parametric component.
\end{abstract}
\section{Introduction}
\IEEEPARstart{R}{obot} platforms are becoming more capable with major advancements in actuator/robot mechanical design. These advancements are evident in the slew of new robotic platforms.
However, one of the limitations present is the inaccuracy of their dynamics models. This inaccuracy may affect many aspects of robots such as; control, stability, motion optimization, and interaction. Any task which is dependent on accurate force control or prediction is subject to issues such as; falling over, crashing into the environment, instability, or other dangerous behaviours. As such, there has been a drive to improve the dynamics models through better measurements and data-driven learning.

Current dynamics models can be split into three major model types: Parametric~\cite{Atkeson1986,Slotine1989,Ting}, Non-Parametric~\cite{Vijayakumar2000,Nguyen-Tuong2009,Cederborg2010}, and Semi-Parametric~\cite{Nguyen-Tuong2010,Camoriano2016,Romeres2016,Riedel2019ComparingRobotics,Romeres2019Derivative-FreeModels}.
Whether the models are Parametric or not depends on the use, or not, of parameters based on known physics. In particular, Parametric models typically use the classical rigid body dynamics equation to model the forces of the robot and heavily rely on reformulations, known as regressors \cite{Atkeson1986,Garofalo2013}, to isolate the inertial dynamic parameters. Parametric models tend to demonstrate strong ability in generalizing over the state space of the robot, with a small enough parameter error and state feedback control. Some examples also benefit from rigorous mathematical proofs of stability \cite{Slotine1989}. The Parametric models do suffer when non-modelled forces are present when implemented on real platforms which may negatively affect performance and learning.

\begin{figure*}
    \vspace*{3mm}
    \captionsetup[subfigure]{aboveskip=0.5\baselineskip,belowskip=0.8\baselineskip}
    \centering
    \begin{subfigure}{0.145\textwidth}
        \includegraphics[width=\linewidth]{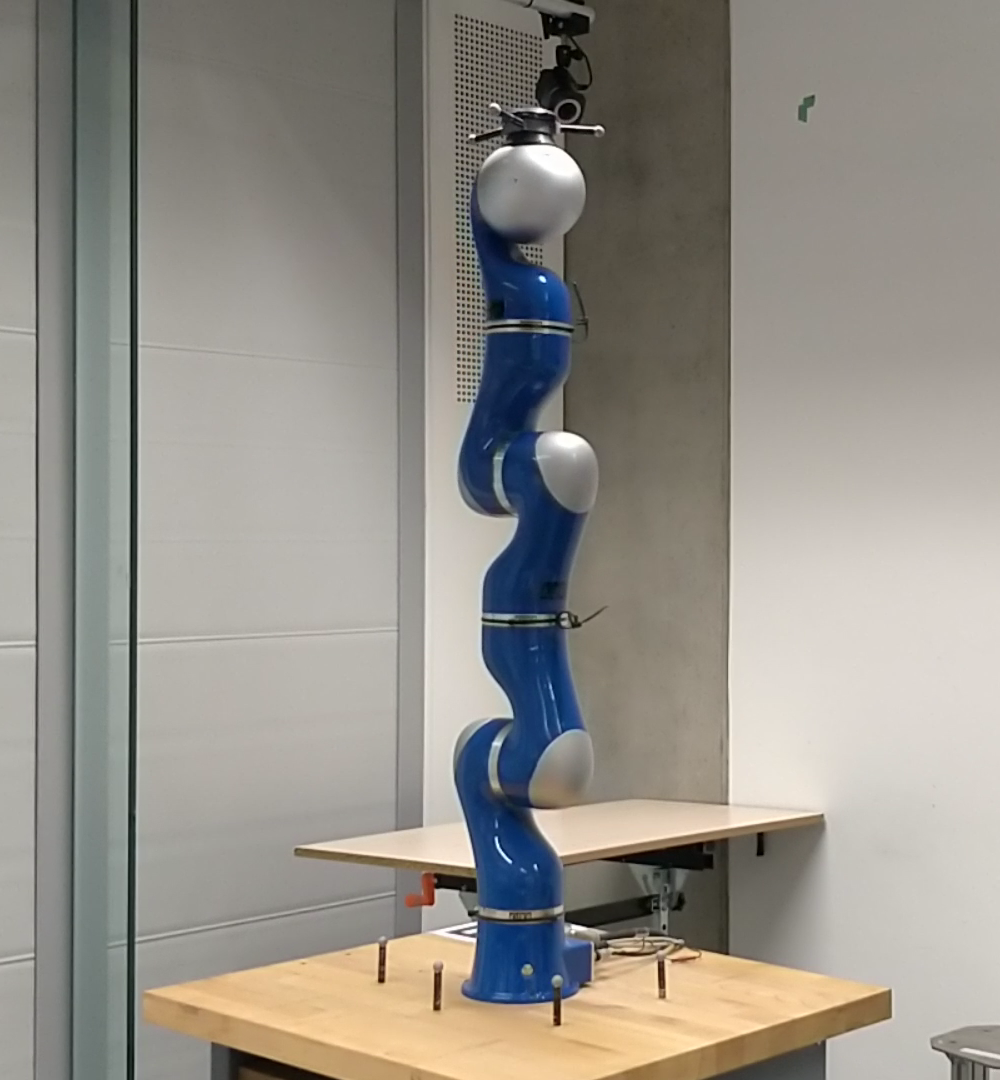}
    \end{subfigure}
    \begin{subfigure}{0.145\textwidth}
        \includegraphics[width=\linewidth]{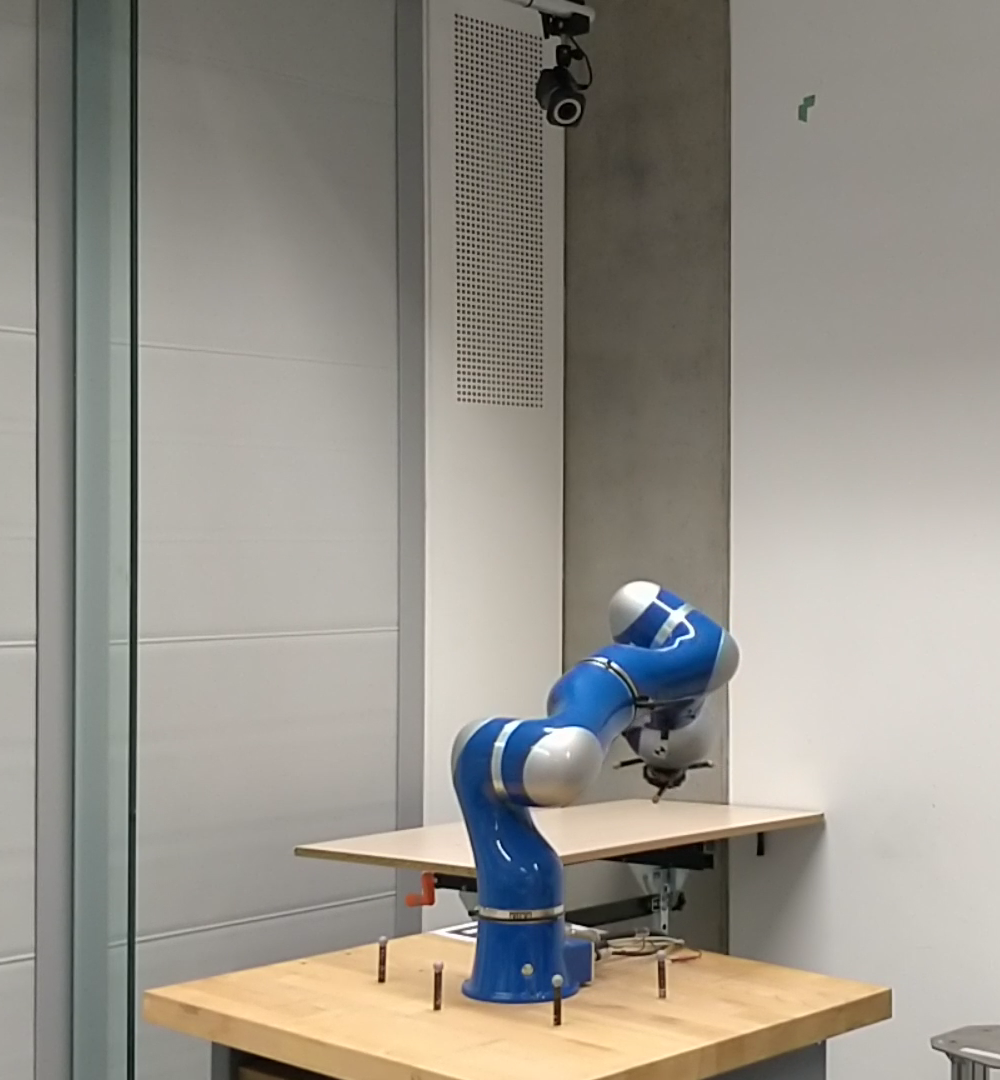}
    \end{subfigure}
    \begin{subfigure}{0.145\textwidth}
        \includegraphics[width=\linewidth]{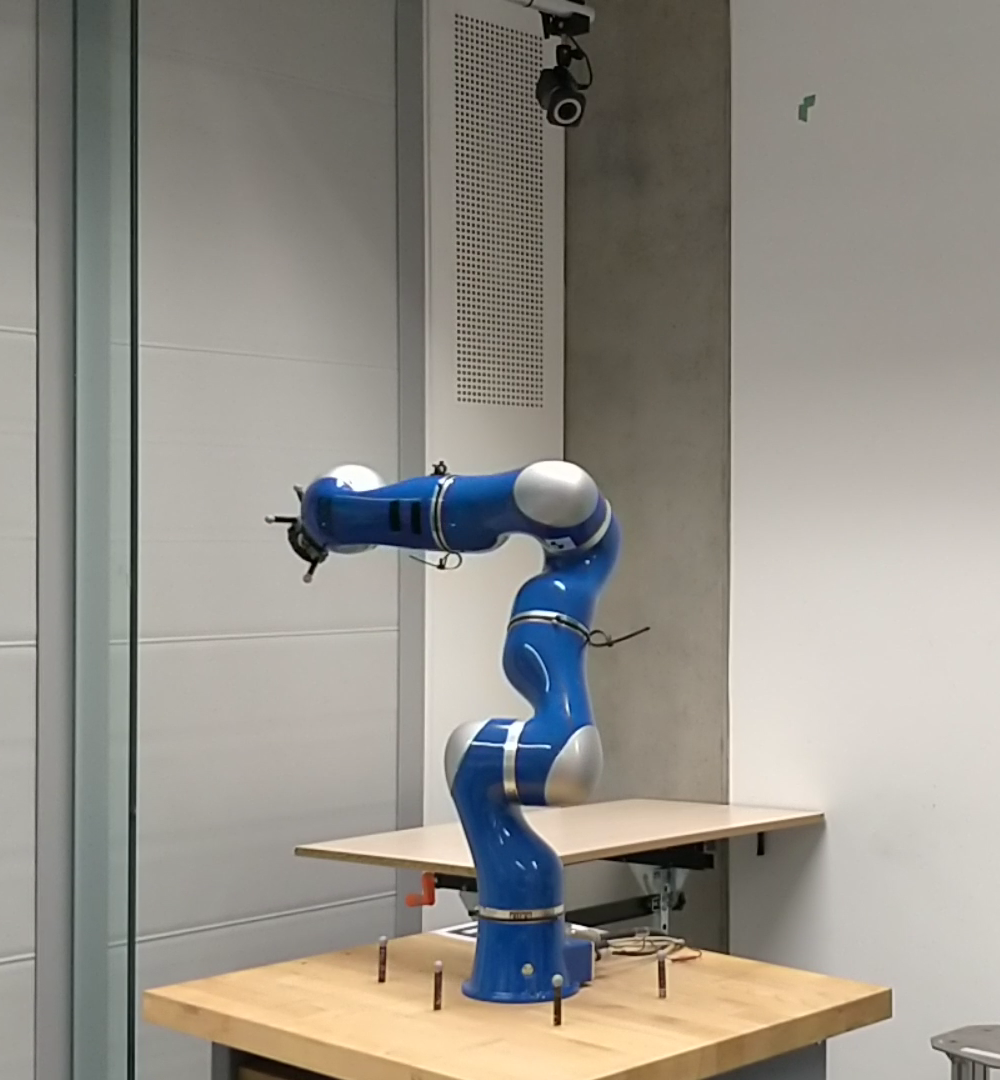}
    \end{subfigure}
    \begin{subfigure}{0.145\textwidth}
        \includegraphics[width=\linewidth]{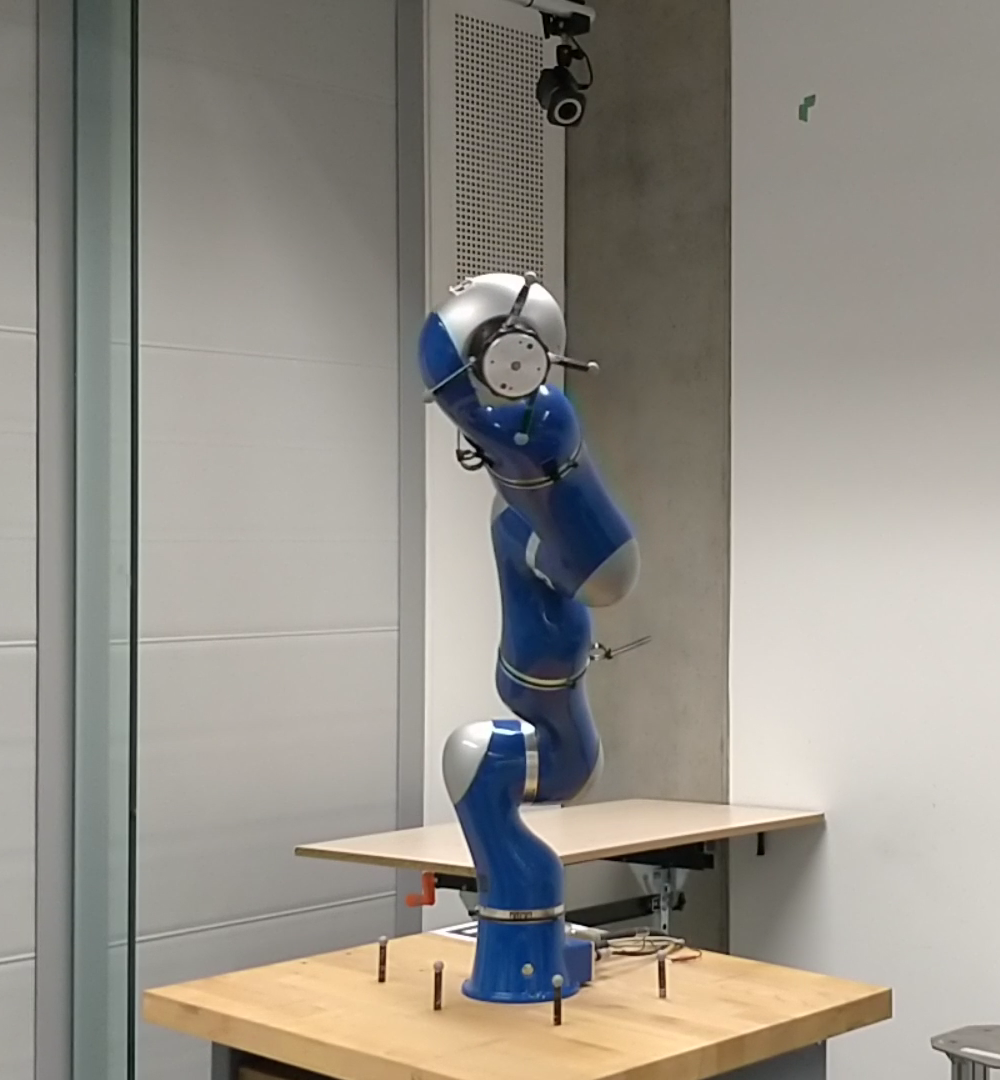}
    \end{subfigure}
    \begin{subfigure}{0.145\textwidth}
        \includegraphics[width=\linewidth]{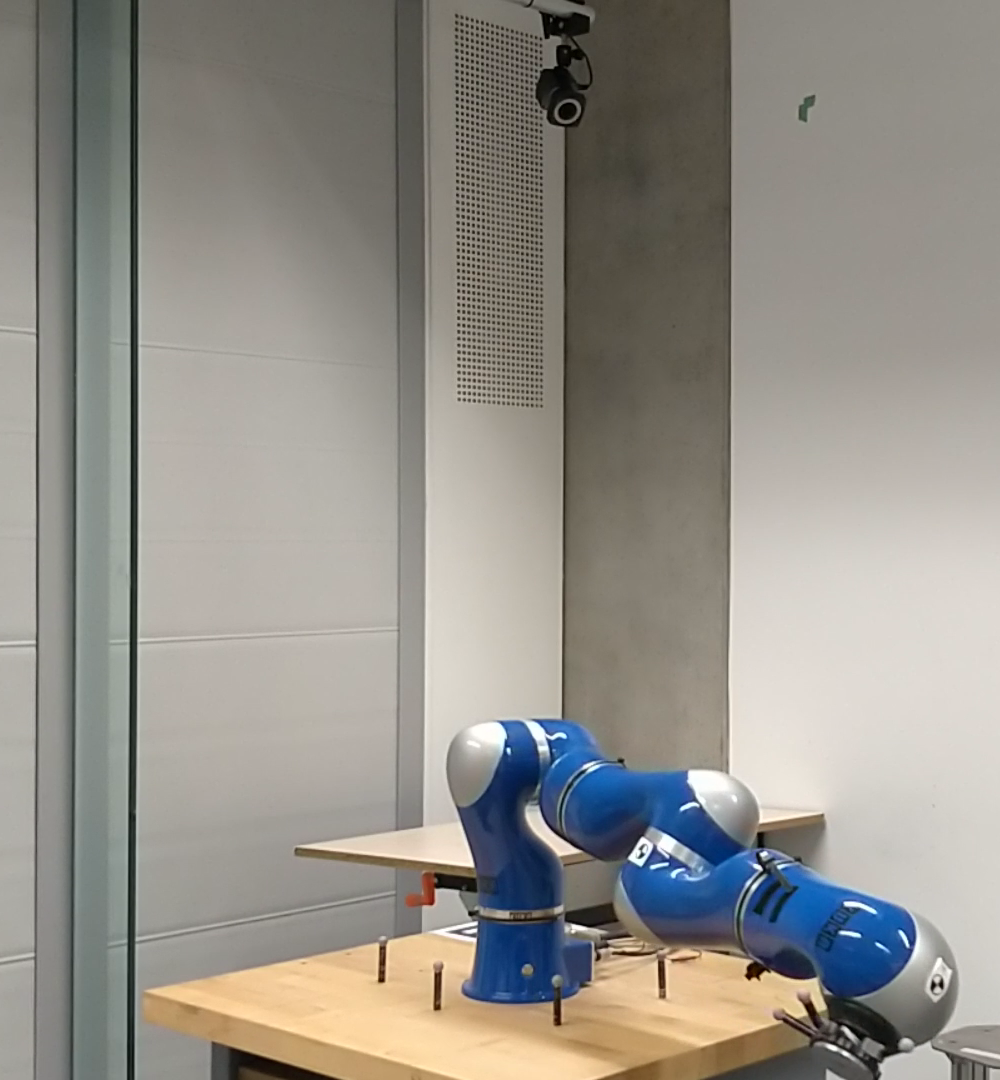}
    \end{subfigure}
    \begin{subfigure}{0.145\textwidth}
        \includegraphics[width=\linewidth]{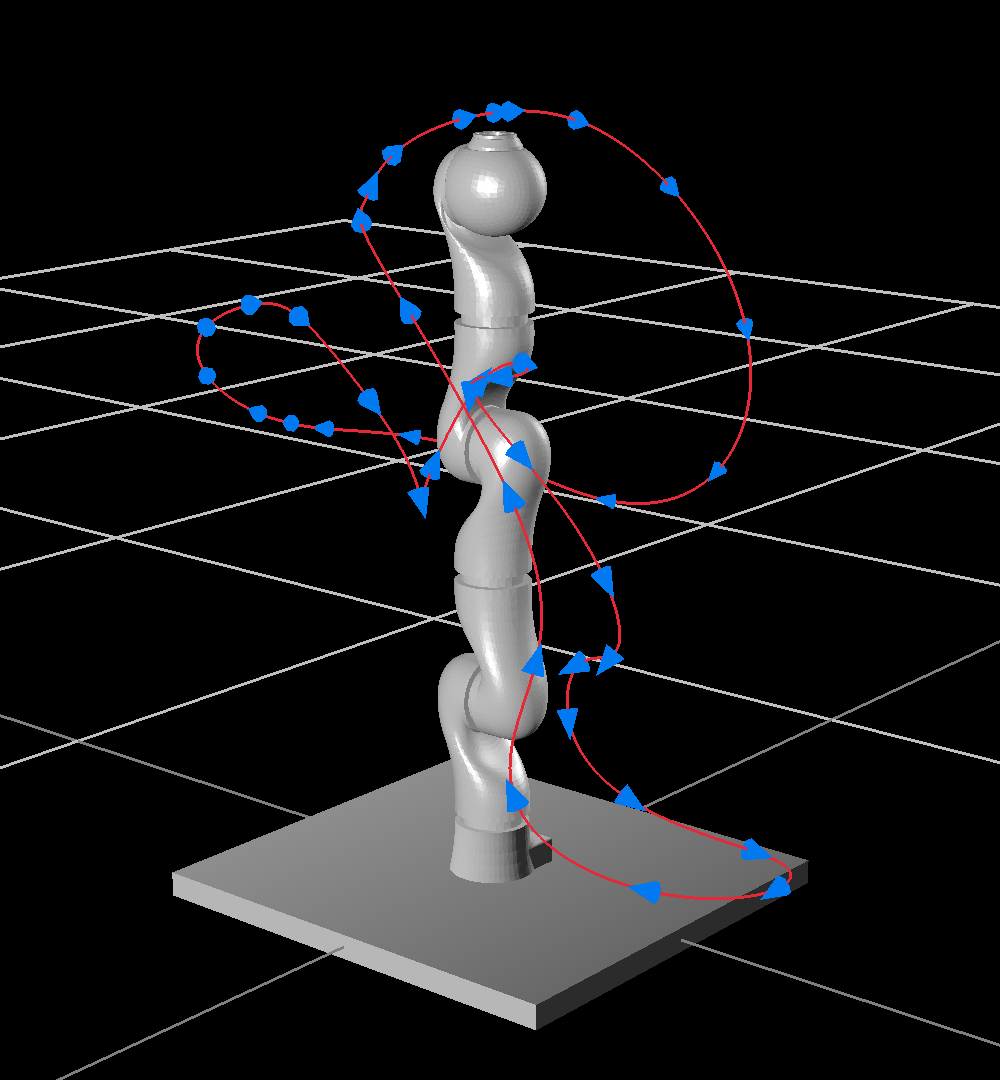}
    \end{subfigure}
    \caption{Snapshots of exemplar Modified Fourier Trajectory with Kuka LWR IV -- The real arm was snapshot during the experiment in \cref{subsec:RPE}, where it is initially controlled only through feedback control as shown in the first 5 figures. These photos provide an idea of the variety of configurations the robot explores. The last figure shows a visualization of the whole trajectory of the end-effector that the arm performs during the experiment.}\label{fig:snaps}
\end{figure*}

Non-Parametric models fall under the context of machine learning, viewing the problem as a mapping function from the input state to the output torque. These models typically do not enforce any strict structure based on physics, allowing the model to learn any non-linear function which can encapsulate all and any forces present in the dynamics of the robot. Non-Parametric methods are normally data-driven as in the case of Gaussian Process Regression~\cite{Nguyen-Tuong2009}, Neural Networks and Gaussian Mixture Models~\cite{Cederborg2010}. The data-driven nature of the algorithms may cause issues if the data is of poor quality or limited in quantity which also affects the generalizability of the model.

Semi-Parametric models are one of the most recent categories proposed to model dynamics and are a combination of both the previous models, Parametric and Non-Parametric. The implementations vary in structure, but generally the Parametric component is designed to handle the traditional rigid body dynamics, whilst the Non-Parametric component designed for a non-linear error which represents all forces not defined in the Parametric component. As an example in \cite{Camoriano2016,Romeres2016} the authors use similar techniques using Random Feature mapping and Recursive Regularized Least Squares to learn the Non-Parametric component whilst using standard least squares for the Parametric component for the Semi-Parametric with rigid body dynamics (RBD) mean with data in small batches. We use the Semi-Parametric model with RBD mean in \cite{Camoriano2016,Romeres2016} as our baseline in \cref{subsec:VME} as it allows independent changes to both components.

The main focus of this paper will be on the issue of online simultaneous model updates. We focus on an online algorithm for faster and continuous learning of the dynamics model. The online aspect will also allow a faster reaction to any physical changes in the dynamics model. Whilst batch-based methods can update with respect to long term physical changes in the dynamics model, they are restricted by the time it takes to collect a batch before applying an update to the model causing higher frequency changes to be missed.

Simultaneously updating both components of the model introduces the issue of a non-stationary target for the Non-Parametric component, which shall be described in the next section. We aim to solve this using a consistency transform that will avoid the need to retrain the Non-Parametric component, or the need to ignore previous data.

The contribution of this work is as follows:
\begin{itemize}
\item Introduction of the non-stationary issue during simultaneous learning for Semi-Parametric models.
\item Proposal of an approximate consistency transform.
\item Defining and analyzing an online simultaneous Semi-Parametric algorithm.
\end{itemize}
\section{Problem Statement}\label{sec:ProblemStatement}
The true dynamics of a fixed robotic platform, with no contact, can be represented as:
\begin{equation}
    \mathbf{M}(q)\ddot{q} + \mathbf{C}(q,\dot{q})\dot{q} + G(q) + F(q,\dot{q}) + \epsilon = \tau
\end{equation}
Where $\mathbf{M}$ represents the inertia matrix, $\mathbf{C}$ the Coriolis and centrifugal matrix, $G$ the gravity vector and $F$ the Coulomb friction vector, which combined form the Parametric contribution of the model. The $\epsilon$ vector represents any other forces present, such as friction effects not in the coulomb model. $\tau$ is the joint torques due to the sum of these forces which are the full torques experienced/measured by the platform.

As mentioned the two components of the Semi-Parametric model learns the full torque model, $\tau$, where $\epsilon$ is learned by the Non-Parametric component which has the following form:
\begin{gather}
    \epsilon = f(q,\dot{q},\ddot{q}) = \tau - \tau_{c}\label{eqn:tauError}\\
    \tau_{c} = \mathbf{M}(q)\ddot{q} + \mathbf{C}(q,\dot{q})\dot{q} + G(q) + F(q,\dot{q})
\end{gather}
Thus $\epsilon$ represents the torque error between the current state's ($q$, $\dot{q}$, $\ddot{q}$) parametric modelled torque, $\tau_{c}$, and the full measured torque output, $\tau$. 

With the formulation of the model in \eqref{eqn:tauError}, when we update the Parametric component the target function of the Non-Parametric component will change. Due to the change in the Non-Parametric target function, the problem is classed as non-stationary with respect to change in the inertial parameters. This non-stationary function means that the previously learned component may be inconsistent with new errors, when the Parametric component is updated, leading to incorrect predictions. The inaccurate predictions can increase errors in trajectory tracking relying on state feedback to control these incorrect torques or lead to instability.

The non-stationary nature of the problem is an important issue if we wish to maximize the contribution of the Parametric component such that the overall model becomes more state generalizable. The same issue can occur if we want to also improve estimations of the inertial parameters through the Parametric component, which can be used by other model predictive controllers. 

To solve the issue of the non-stationary target function we propose an online Semi-Parametric algorithm with a consistency transform for the Non-Parametric component with respect to the known inertial parameter change. The consistency transform will allow the previously learned component to approximately remain consistent with the new target function without further learning or retraining.
\section{Methodology}\label{sec:Methodology}
\begin{figure*}
  \centering
  \begin{subfigure}{0.3\linewidth}
      \centering
          
\begingroup
\makeatletter
\providecommand\color[2][]{%
  \GenericError{(gnuplot) \space\space\space\@spaces}{%
    Package color not loaded in conjunction with
    terminal option `colourtext'%
  }{See the gnuplot documentation for explanation.%
  }{Either use 'blacktext' in gnuplot or load the package
    color.sty in LaTeX.}%
  \renewcommand\color[2][]{}%
}%
\providecommand\includegraphics[2][]{%
  \GenericError{(gnuplot) \space\space\space\@spaces}{%
    Package graphicx or graphics not loaded%
  }{See the gnuplot documentation for explanation.%
  }{The gnuplot epslatex terminal needs graphicx.sty or graphics.sty.}%
  \renewcommand\includegraphics[2][]{}%
}%
\providecommand\rotatebox[2]{#2}%
\@ifundefined{ifGPcolor}{%
  \newif\ifGPcolor
  \GPcolortrue
}{}%
\@ifundefined{ifGPblacktext}{%
  \newif\ifGPblacktext
  \GPblacktexttrue
}{}%
\let\gplgaddtomacro\g@addto@macro
\gdef\gplbacktext{}%
\gdef\gplfronttext{}%
\makeatother
\ifGPblacktext
  \def\colorrgb#1{}%
  \def\colorgray#1{}%
\else
  \ifGPcolor
    \def\colorrgb#1{\color[rgb]{#1}}%
    \def\colorgray#1{\color[gray]{#1}}%
    \expandafter\def\csname LTw\endcsname{\color{white}}%
    \expandafter\def\csname LTb\endcsname{\color{black}}%
    \expandafter\def\csname LTa\endcsname{\color{black}}%
    \expandafter\def\csname LT0\endcsname{\color[rgb]{1,0,0}}%
    \expandafter\def\csname LT1\endcsname{\color[rgb]{0,1,0}}%
    \expandafter\def\csname LT2\endcsname{\color[rgb]{0,0,1}}%
    \expandafter\def\csname LT3\endcsname{\color[rgb]{1,0,1}}%
    \expandafter\def\csname LT4\endcsname{\color[rgb]{0,1,1}}%
    \expandafter\def\csname LT5\endcsname{\color[rgb]{1,1,0}}%
    \expandafter\def\csname LT6\endcsname{\color[rgb]{0,0,0}}%
    \expandafter\def\csname LT7\endcsname{\color[rgb]{1,0.3,0}}%
    \expandafter\def\csname LT8\endcsname{\color[rgb]{0.5,0.5,0.5}}%
  \else
    \def\colorrgb#1{\color{black}}%
    \def\colorgray#1{\color[gray]{#1}}%
    \expandafter\def\csname LTw\endcsname{\color{white}}%
    \expandafter\def\csname LTb\endcsname{\color{black}}%
    \expandafter\def\csname LTa\endcsname{\color{black}}%
    \expandafter\def\csname LT0\endcsname{\color{black}}%
    \expandafter\def\csname LT1\endcsname{\color{black}}%
    \expandafter\def\csname LT2\endcsname{\color{black}}%
    \expandafter\def\csname LT3\endcsname{\color{black}}%
    \expandafter\def\csname LT4\endcsname{\color{black}}%
    \expandafter\def\csname LT5\endcsname{\color{black}}%
    \expandafter\def\csname LT6\endcsname{\color{black}}%
    \expandafter\def\csname LT7\endcsname{\color{black}}%
    \expandafter\def\csname LT8\endcsname{\color{black}}%
  \fi
\fi
  \setlength{\unitlength}{0.0500bp}%
  \ifx\gptboxheight\undefined%
    \newlength{\gptboxheight}%
    \newlength{\gptboxwidth}%
    \newsavebox{\gptboxtext}%
  \fi%
  \setlength{\fboxrule}{0.5pt}%
  \setlength{\fboxsep}{1pt}%
\begin{picture}(3020.00,1500.00)%
  \gplgaddtomacro\gplbacktext{%
    \csname LTb\endcsname
    \put(357,372){\makebox(0,0)[r]{\strut{}$-2$}}%
    \csname LTb\endcsname
    \put(357,607){\makebox(0,0)[r]{\strut{}$-1$}}%
    \csname LTb\endcsname
    \put(357,843){\makebox(0,0)[r]{\strut{}$0$}}%
    \csname LTb\endcsname
    \put(357,1078){\makebox(0,0)[r]{\strut{}$1$}}%
    \csname LTb\endcsname
    \put(357,1313){\makebox(0,0)[r]{\strut{}$2$}}%
    \csname LTb\endcsname
    \put(459,186){\makebox(0,0){\strut{}$-1$}}%
    \csname LTb\endcsname
    \put(709,186){\makebox(0,0){\strut{}$0$}}%
    \csname LTb\endcsname
    \put(960,186){\makebox(0,0){\strut{}$1$}}%
    \csname LTb\endcsname
    \put(1210,186){\makebox(0,0){\strut{}$2$}}%
    \csname LTb\endcsname
    \put(1461,186){\makebox(0,0){\strut{}$3$}}%
    \csname LTb\endcsname
    \put(1711,186){\makebox(0,0){\strut{}$4$}}%
    \csname LTb\endcsname
    \put(1962,186){\makebox(0,0){\strut{}$5$}}%
    \csname LTb\endcsname
    \put(2212,186){\makebox(0,0){\strut{}$6$}}%
    \csname LTb\endcsname
    \put(2463,186){\makebox(0,0){\strut{}$7$}}%
    \csname LTb\endcsname
    \put(2713,186){\makebox(0,0){\strut{}$8$}}%
  }%
  \gplgaddtomacro\gplfronttext{%
    \csname LTb\endcsname
    \put(2231,1146){\makebox(0,0)[r]{\strut{}sin(x)}}%
  }%
  \gplbacktext
  \put(0,0){\includegraphics{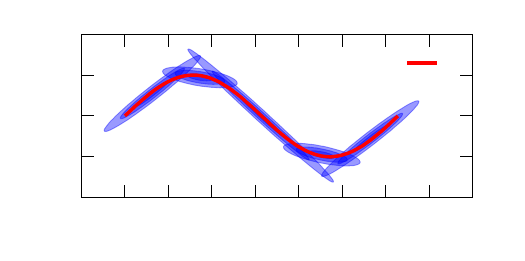}}%
  \gplfronttext
\end{picture}%
\endgroup

      \vspace*{-0.8\baselineskip}
      \caption{}\label{fig:GMMSine}
  \end{subfigure}
  \begin{subfigure}{0.3\linewidth}
  
      \centering

          
\begingroup
\makeatletter
\providecommand\color[2][]{%
  \GenericError{(gnuplot) \space\space\space\@spaces}{%
    Package color not loaded in conjunction with
    terminal option `colourtext'%
  }{See the gnuplot documentation for explanation.%
  }{Either use 'blacktext' in gnuplot or load the package
    color.sty in LaTeX.}%
  \renewcommand\color[2][]{}%
}%
\providecommand\includegraphics[2][]{%
  \GenericError{(gnuplot) \space\space\space\@spaces}{%
    Package graphicx or graphics not loaded%
  }{See the gnuplot documentation for explanation.%
  }{The gnuplot epslatex terminal needs graphicx.sty or graphics.sty.}%
  \renewcommand\includegraphics[2][]{}%
}%
\providecommand\rotatebox[2]{#2}%
\@ifundefined{ifGPcolor}{%
  \newif\ifGPcolor
  \GPcolortrue
}{}%
\@ifundefined{ifGPblacktext}{%
  \newif\ifGPblacktext
  \GPblacktexttrue
}{}%
\let\gplgaddtomacro\g@addto@macro
\gdef\gplbacktext{}%
\gdef\gplfronttext{}%
\makeatother
\ifGPblacktext
  \def\colorrgb#1{}%
  \def\colorgray#1{}%
\else
  \ifGPcolor
    \def\colorrgb#1{\color[rgb]{#1}}%
    \def\colorgray#1{\color[gray]{#1}}%
    \expandafter\def\csname LTw\endcsname{\color{white}}%
    \expandafter\def\csname LTb\endcsname{\color{black}}%
    \expandafter\def\csname LTa\endcsname{\color{black}}%
    \expandafter\def\csname LT0\endcsname{\color[rgb]{1,0,0}}%
    \expandafter\def\csname LT1\endcsname{\color[rgb]{0,1,0}}%
    \expandafter\def\csname LT2\endcsname{\color[rgb]{0,0,1}}%
    \expandafter\def\csname LT3\endcsname{\color[rgb]{1,0,1}}%
    \expandafter\def\csname LT4\endcsname{\color[rgb]{0,1,1}}%
    \expandafter\def\csname LT5\endcsname{\color[rgb]{1,1,0}}%
    \expandafter\def\csname LT6\endcsname{\color[rgb]{0,0,0}}%
    \expandafter\def\csname LT7\endcsname{\color[rgb]{1,0.3,0}}%
    \expandafter\def\csname LT8\endcsname{\color[rgb]{0.5,0.5,0.5}}%
  \else
    \def\colorrgb#1{\color{black}}%
    \def\colorgray#1{\color[gray]{#1}}%
    \expandafter\def\csname LTw\endcsname{\color{white}}%
    \expandafter\def\csname LTb\endcsname{\color{black}}%
    \expandafter\def\csname LTa\endcsname{\color{black}}%
    \expandafter\def\csname LT0\endcsname{\color{black}}%
    \expandafter\def\csname LT1\endcsname{\color{black}}%
    \expandafter\def\csname LT2\endcsname{\color{black}}%
    \expandafter\def\csname LT3\endcsname{\color{black}}%
    \expandafter\def\csname LT4\endcsname{\color{black}}%
    \expandafter\def\csname LT5\endcsname{\color{black}}%
    \expandafter\def\csname LT6\endcsname{\color{black}}%
    \expandafter\def\csname LT7\endcsname{\color{black}}%
    \expandafter\def\csname LT8\endcsname{\color{black}}%
  \fi
\fi
  \setlength{\unitlength}{0.0500bp}%
  \ifx\gptboxheight\undefined%
    \newlength{\gptboxheight}%
    \newlength{\gptboxwidth}%
    \newsavebox{\gptboxtext}%
  \fi%
  \setlength{\fboxrule}{0.5pt}%
  \setlength{\fboxsep}{1pt}%
\begin{picture}(3020.00,1500.00)%
  \gplgaddtomacro\gplbacktext{%
    \csname LTb\endcsname
    \put(357,372){\makebox(0,0)[r]{\strut{}$-2$}}%
    \csname LTb\endcsname
    \put(357,607){\makebox(0,0)[r]{\strut{}$-1$}}%
    \csname LTb\endcsname
    \put(357,843){\makebox(0,0)[r]{\strut{}$0$}}%
    \csname LTb\endcsname
    \put(357,1078){\makebox(0,0)[r]{\strut{}$1$}}%
    \csname LTb\endcsname
    \put(357,1313){\makebox(0,0)[r]{\strut{}$2$}}%
    \csname LTb\endcsname
    \put(459,186){\makebox(0,0){\strut{}$-1$}}%
    \csname LTb\endcsname
    \put(709,186){\makebox(0,0){\strut{}$0$}}%
    \csname LTb\endcsname
    \put(960,186){\makebox(0,0){\strut{}$1$}}%
    \csname LTb\endcsname
    \put(1210,186){\makebox(0,0){\strut{}$2$}}%
    \csname LTb\endcsname
    \put(1461,186){\makebox(0,0){\strut{}$3$}}%
    \csname LTb\endcsname
    \put(1711,186){\makebox(0,0){\strut{}$4$}}%
    \csname LTb\endcsname
    \put(1962,186){\makebox(0,0){\strut{}$5$}}%
    \csname LTb\endcsname
    \put(2212,186){\makebox(0,0){\strut{}$6$}}%
    \csname LTb\endcsname
    \put(2463,186){\makebox(0,0){\strut{}$7$}}%
    \csname LTb\endcsname
    \put(2713,186){\makebox(0,0){\strut{}$8$}}%
  }%
  \gplgaddtomacro\gplfronttext{%
    \csname LTb\endcsname
    \put(2231,1146){\makebox(0,0)[r]{\strut{}0.2 sin(x)}}%
  }%
  \gplbacktext
  \put(0,0){\includegraphics{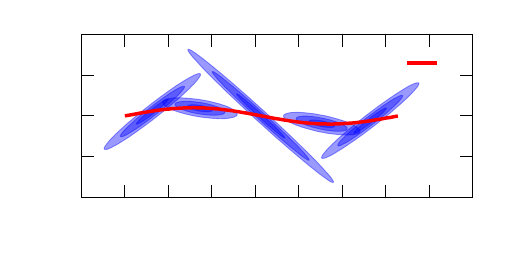}}%
  \gplfronttext
\end{picture}%
\endgroup

      \vspace*{-0.8\baselineskip}
      \caption{}\label{fig:GMMSineT}
  \end{subfigure}
  \begin{subfigure}{0.3\linewidth}
  
      \centering
          
\begingroup
\makeatletter
\providecommand\color[2][]{%
  \GenericError{(gnuplot) \space\space\space\@spaces}{%
    Package color not loaded in conjunction with
    terminal option `colourtext'%
  }{See the gnuplot documentation for explanation.%
  }{Either use 'blacktext' in gnuplot or load the package
    color.sty in LaTeX.}%
  \renewcommand\color[2][]{}%
}%
\providecommand\includegraphics[2][]{%
  \GenericError{(gnuplot) \space\space\space\@spaces}{%
    Package graphicx or graphics not loaded%
  }{See the gnuplot documentation for explanation.%
  }{The gnuplot epslatex terminal needs graphicx.sty or graphics.sty.}%
  \renewcommand\includegraphics[2][]{}%
}%
\providecommand\rotatebox[2]{#2}%
\@ifundefined{ifGPcolor}{%
  \newif\ifGPcolor
  \GPcolortrue
}{}%
\@ifundefined{ifGPblacktext}{%
  \newif\ifGPblacktext
  \GPblacktexttrue
}{}%
\let\gplgaddtomacro\g@addto@macro
\gdef\gplbacktext{}%
\gdef\gplfronttext{}%
\makeatother
\ifGPblacktext
  \def\colorrgb#1{}%
  \def\colorgray#1{}%
\else
  \ifGPcolor
    \def\colorrgb#1{\color[rgb]{#1}}%
    \def\colorgray#1{\color[gray]{#1}}%
    \expandafter\def\csname LTw\endcsname{\color{white}}%
    \expandafter\def\csname LTb\endcsname{\color{black}}%
    \expandafter\def\csname LTa\endcsname{\color{black}}%
    \expandafter\def\csname LT0\endcsname{\color[rgb]{1,0,0}}%
    \expandafter\def\csname LT1\endcsname{\color[rgb]{0,1,0}}%
    \expandafter\def\csname LT2\endcsname{\color[rgb]{0,0,1}}%
    \expandafter\def\csname LT3\endcsname{\color[rgb]{1,0,1}}%
    \expandafter\def\csname LT4\endcsname{\color[rgb]{0,1,1}}%
    \expandafter\def\csname LT5\endcsname{\color[rgb]{1,1,0}}%
    \expandafter\def\csname LT6\endcsname{\color[rgb]{0,0,0}}%
    \expandafter\def\csname LT7\endcsname{\color[rgb]{1,0.3,0}}%
    \expandafter\def\csname LT8\endcsname{\color[rgb]{0.5,0.5,0.5}}%
  \else
    \def\colorrgb#1{\color{black}}%
    \def\colorgray#1{\color[gray]{#1}}%
    \expandafter\def\csname LTw\endcsname{\color{white}}%
    \expandafter\def\csname LTb\endcsname{\color{black}}%
    \expandafter\def\csname LTa\endcsname{\color{black}}%
    \expandafter\def\csname LT0\endcsname{\color{black}}%
    \expandafter\def\csname LT1\endcsname{\color{black}}%
    \expandafter\def\csname LT2\endcsname{\color{black}}%
    \expandafter\def\csname LT3\endcsname{\color{black}}%
    \expandafter\def\csname LT4\endcsname{\color{black}}%
    \expandafter\def\csname LT5\endcsname{\color{black}}%
    \expandafter\def\csname LT6\endcsname{\color{black}}%
    \expandafter\def\csname LT7\endcsname{\color{black}}%
    \expandafter\def\csname LT8\endcsname{\color{black}}%
  \fi
\fi
  \setlength{\unitlength}{0.0500bp}%
  \ifx\gptboxheight\undefined%
    \newlength{\gptboxheight}%
    \newlength{\gptboxwidth}%
    \newsavebox{\gptboxtext}%
  \fi%
  \setlength{\fboxrule}{0.5pt}%
  \setlength{\fboxsep}{1pt}%
\begin{picture}(3020.00,1500.00)%
  \gplgaddtomacro\gplbacktext{%
    \csname LTb\endcsname
    \put(357,372){\makebox(0,0)[r]{\strut{}$-2$}}%
    \csname LTb\endcsname
    \put(357,607){\makebox(0,0)[r]{\strut{}$-1$}}%
    \csname LTb\endcsname
    \put(357,843){\makebox(0,0)[r]{\strut{}$0$}}%
    \csname LTb\endcsname
    \put(357,1078){\makebox(0,0)[r]{\strut{}$1$}}%
    \csname LTb\endcsname
    \put(357,1313){\makebox(0,0)[r]{\strut{}$2$}}%
    \csname LTb\endcsname
    \put(459,186){\makebox(0,0){\strut{}$-1$}}%
    \csname LTb\endcsname
    \put(741,186){\makebox(0,0){\strut{}$0$}}%
    \csname LTb\endcsname
    \put(1023,186){\makebox(0,0){\strut{}$1$}}%
    \csname LTb\endcsname
    \put(1304,186){\makebox(0,0){\strut{}$2$}}%
    \csname LTb\endcsname
    \put(1586,186){\makebox(0,0){\strut{}$3$}}%
    \csname LTb\endcsname
    \put(1868,186){\makebox(0,0){\strut{}$4$}}%
    \csname LTb\endcsname
    \put(2150,186){\makebox(0,0){\strut{}$5$}}%
    \csname LTb\endcsname
    \put(2431,186){\makebox(0,0){\strut{}$6$}}%
    \csname LTb\endcsname
    \put(2713,186){\makebox(0,0){\strut{}$7$}}%
  }%
  \gplgaddtomacro\gplfronttext{%
    \csname LTb\endcsname
    \put(2231,1146){\makebox(0,0)[r]{\strut{}0.2 sin(x)}}%
  }%
  \gplbacktext
  \put(0,0){\includegraphics{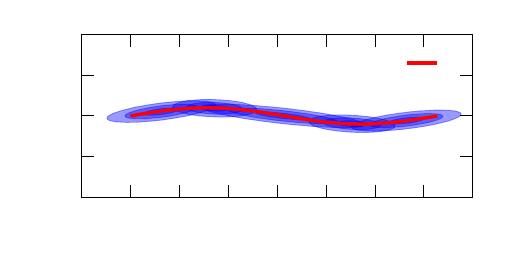}}%
  \gplfronttext
\end{picture}%
\endgroup

      \vspace*{-0.8\baselineskip}
      \caption{}\label{fig:GMMSineTSig}
  \end{subfigure}
    \caption{Consistency Example - Starting from the original Sine Wave (a), we demonstrate the idea of transforming the GMM with a known change in the local function. In this example, we take a standard Sine Wave and multiply it by 0.2 and apply the transformations from \cref{sec:SPConsis}. (b) shows the mean transformation of the GMM sub-models which put the centre of each Gaussian onto the new updated function. (c) shows the result when the covariance transformation is applied to the sub-models in (b). The covariances adjust to the new slopes of the function and the new GMM is now consistent with the change in the function. It is important to note that no learning is done between (a) - (c), only the consistency transform applied.}\label{fig:consis}
    \end{figure*}
In the implementation of this paper, we have used Composite Adaptive Control~\cite{Slotine1989} and Gaussian Mixture Models~\cite{Pinto2017ScalableModels} (GMMs) for the Parametric and Non-Parametric components respectively. The Parametric component can be freely chosen, however,~\cite{Slotine1989} was chosen due to the proven stable adaptation. The Composite Adaptive Control algorithm also defines two sources of learning, the state (position and velocity) and torque errors. The two errors can be beneficial for Semi-Parametric controllers, as often the state errors are driven towards zero quickly, due to both components learning the correct torque combined. When the state errors approach zero the learning using this error stalls, meaning that the Parametric component will stop learning even if the inertial parameters are incorrect. By using the torque errors as well we can still drive the Parametric learning even if the state errors tend towards zero, allowing us to maximize the Parametric component's contribution and to minimize the error on the inertial parameters.

GMMs were chosen for the Non-Parametric component for several reasons. Firstly, efficient implementations of the algorithm exist as evidenced in~\cite{Pinto2017ScalableModels}, which compared against other methods such as Gaussian Process Regression, have a significantly smaller complexity. Secondly, the representation is statistically significant, which can reduce the amount of memory required due to previous data being described by the statistics. The last reason is due to the GMM's statistical nature, the space is easier to understand allowing a linear transformation to be defined and applied.

\subsection{Parametric Model}\label{sec:Parametric}
The basic approach for the Parametric learning component is to rearrange the rigid body dynamics equations to be linear with respect to the dynamic parameters, creating a non-linear matrix based on the robot state and the kinematic parameters of the robot known as a regressor. The regressor is computable in closed form given kinematic knowledge of the robot platform. The regressor can then be inverted or used in an iterative least-squares algorithm to find the inertial parameters given input data. In this work we use a Composite Adaptive Control scheme~\cite{Slotine1989}, this combines both dynamics learning and control in a way that is provably stable with convergence in position, velocity and inertial parameters.

Composite Adaptive Control~\cite{Slotine1989} learns using two sources of error from two different regressors. These regressors are defined as direct and indirect in~\cite{Slotine1989}. The direct regressor formulation uses \eqref{eqn:er}, known as resolved accelerations and velocities, to eliminate the need for acceleration measurements by using the knowledge of the desired trajectory. This allows the model learning to be driven by the position and velocity errors during the task, which can be used to show stable convergence in the tracking errors. We can then reformulate the rigid body dynamics equation into a matrix which is independent of the inertial parameters \eqref{eqn:Ysol}.
\begin{align}
    \dot{q}_r = \dot{q}_d - \mathbold{{\Lambda}} (q-q_d) & &  \ddot{q}_r = \ddot{q}_d - \mathbold{{\Lambda}} (\dot{q}-\dot{q}_d)\label{eqn:er}
\end{align}
\begin{gather}
    \mathbf{M}(q)\ddot{q}_{r} + \mathbf{C}(q,\dot{q})\dot{q}_{r} + G(q) + F(q,\dot{q}) = \mathbf{Y}(q,\dot{q},\dot{q}_{r},\ddot{q}_{r})\pi\label{eqn:Ysol}
\end{gather}
Where $\bullet_d$ is the desired value of $\bullet$, $\mathbold{{\Lambda}}$ is a positive definite gain matrix, $\mathbf{Y}$ is the direct regressor, and $\pi$ is the inertial parameter vector. Typically for each link $i$ the inertial parameters are: $\pi_{i} = [m_{i}, m_{i}{c_{i}}^{T}, {I_{xx}}_{i}, {I_{xy}}_{i}, {I_{xz}}_{i}, {I_{yy}}_{i}, {I_{yz}}_{i}, {I_{zz}}_{i}]^T$, where $m_{i}$, $c_{i}$ and $I_{i}$ are the mass, centre of mass and link inertia respectively.

The direct regressor suffers with parameter convergence of the model as the learning is not dependent on the torque predictions. The indirect formulation solves this issue by driving the update law on the predicted filtered torque error at the given state of the robot. The indirect regressor uses a filtering approach to reformulate the dynamics equation without the need for acceleration measurements.
\begin{equation}
    \int^{t}_{0}w(r)\tau(r) dr = \mathbf{W}(\dot{q},q)\pi
\end{equation}
Where $w(r)$ is the impulse response from a first-order filter and $\mathbf{W}$ is the filtered regressor.

By combining the two regressors into the single update law \eqref{eqn:SlotineUpdateLaw}, the learning process is driven by both tracking and prediction errors. When combined with the control law in~\cite{Slotine1989}, the controller can be shown to be stable in the Lyapunov sense with convergence in the parameters, velocity and position~\cite{Slotine1989}.
\begin{equation}
    \Delta{\hat{\pi}} = -\mathbf{P}(\mathbf{Y}(q,\dot{q},\dot{q}_{r},\ddot{q}_{r})^T s + \mathbf{W}(q,\dot{q})^T \mathbf{R}e)
    \label{eqn:SlotineUpdateLaw}
\end{equation}
Where $\Delta{\bullet}$ is the change in $\bullet$, $\hat{\bullet}$ is the estimated value of $\bullet$, $s = \dot{q}-\dot{q}_r$, $e$ is the measured torque error, and $\mathbf{P}$ and $\mathbf{R}$ are positive definite weight matrices.

\subsection{Non-Parametric Model}\label{sec:NonParametric}
The general concept of the Gaussian Mixture Model is to represent a non-linear function through a sum of multidimensional Gaussian sub-models. This represents the full joint probability space for the inputs and outputs, as each Gaussian has the dimension of the input plus output dimensions. To define these Gaussian sub-models, to learn a function, they need to be placed throughout the state space following the training data. An example of a trained GMM on a Sine Wave is shown in \cref{fig:GMMSine}.

The GMM learning was implemented using the Iterative Gaussian Mixture Model (IGMM) algorithm defined in~\cite{Pinto2017ScalableModels} which can be learned iteratively at a low computational cost. The algorithm also defines a metric for adding and removing components.

Using the GMM we apply a regression algorithm, known as Gaussian Mixture Regression, to predict an output based on a given input. The regression uses the fact that each Gaussian is defined as a prior, a mean vector ($\mu$), and a covariance matrix ($\Sigma$). As each Gaussian also represents the joint probability of the inputs and outputs, we can decompose the parameters representing the mean and covariance as \eqref{eqn:decomp}. Using this structure we condition each Gaussian on the input dimensions, with a given input, to create a conditional mean for the output dimensions \eqref{eqn:GMR}.
\begin{align}
\mu^{j} = \begin{bmatrix}\mu^{j}_{i}\\\mu^j_{o}\end{bmatrix} & & \Sigma^{j} = \begin{bmatrix}\Sigma^j_{i,i} & \Sigma^j_{i,o}\\ \Sigma^j_{o,i} & \Sigma^j_{o,o}\end{bmatrix}\label{eqn:decomp}
\end{align}
\begin{equation}
    \hat{x}_o = \sum^{M}_{j=1} p\left(j|x_i\right)\left(\mu^j_{o} + \Sigma^{j}_{o,i} {\Sigma^j_{i,i}}^{-1}\left( x_{i} - \mu^j_i \right)\right)\label{eqn:GMR}
\end{equation}
Where the $\bullet^j$ indicates the Gaussian $\bullet$ belongs to, $p(j|x_i)$ is the posterior probability of the input given the $j$th Gaussian, $i$ and $o$ represent the input ($q$, $\dot{q}$, $\ddot{q}$)  and output ($\tau$) dimensions respectively, $\mu_{i}$ and $\mu_{o}$ represent the input and output dimension means, $\Sigma_{k,l}$ represents the covariance between the dimensions $k$ and $l$, and $\hat{x}_o$ represents the conditioned prediction output of the GMR algorithm.

\subsection{Semi-Parametric Model}
The Semi-Parametric model takes both of these components and combines them such that the Parametric component describes the forces related to the dynamic parameters ($\mathbf{M}(q)$,~$\mathbf{C}(q,\dot{q})$,~$G(q)$,~$F(q,\dot{q})$), whilst the Non-Parametric target is the error between the actual torque and the Parametric component. As mentioned in section \ref{sec:ProblemStatement} this can create an inconsistency between the components if both are updated, which could potentially cause erratic or dangerous behaviour through poor predictions.
\subsection{Semi-Parametric Consistency}\label{sec:SPConsis}%
When the Parametric component updates are known with respect to the inertial parameters, we can determine the change in torque in any given state given the change in parameters.  In the following, we demonstrate the linear transformation in the Non-Parametric function with respect to the inertial parameters.
\begin{gather}
    \mathbf{M}(q)\ddot{q} + \mathbf{C}(q,\dot{q})\dot{q} + G(q) + F(q,\dot{q}) + f(q,\dot{q},\ddot{q}) = \tau \label{eqn:SP}\\
    f(q,\dot{q},\ddot{q}) = \tau - \mathbf{Y}(q,\dot{q},\ddot{q})\hat{\pi}  \label{eqn:NPReg}\\
    \frac{\partial{f(q,\dot{q},\ddot{q})}}{\partial \pi} = -\mathbf{Y}(q,\dot{q},\ddot{q})  \label{eqn:NPpi}\\
    \Delta{f}(q,\dot{q},\ddot{q}) = -\mathbf{Y}(q,\dot{q},\ddot{q})\Delta{\hat{\pi}}  \label{eqn:NPpidot}
\end{gather}
Equation \eqref{eqn:NPReg} is obtained through substituting \eqref{eqn:Ysol} in \eqref{eqn:SP}. Then taking the derivative with respect to the inertial parameters ($\pi$) results in \eqref{eqn:NPpi}. When multiplying this by the change in inertial parameters ($\Delta{\pi}$) leads to \eqref{eqn:NPpidot}. 
Using \eqref{eqn:NPpidot} we can approximately update each Gaussian in the GMM in their local space by linearly approximating the change in torque with respect to state and parameter update. In particular noting that each of the Gaussian sub-models is described through two main variables; the mean, and the covariance, which we can update with linearly approximated space transformations. \Cref{fig:consis} gives a visual example of the transformation.

The mean is particularly straightforward to update using:
\begin{equation}
  \tau_{\mu^{\prime}} = \tau_{\mu} -\mathbf{Y}(q_{\mu},\dot{q}_{\mu},\ddot{q}_{\mu})\Delta{\hat{\pi}}
\end{equation}
Where $\bullet_{\mu}$ is the metric $\bullet$ obtained from the relevant dimensions of $\mu$ of the Gaussian and $\bullet^{\prime}$ is the updated metric. 

The covariance is not as straightforward as it lies across the relevant space that changes non-linearly. We should be able to make a rough assumption that the covariance matrix in this situation roughly is equivalent to \eqref{eqn:covarianceLayout} for the case of dynamics.

\begin{equation}\label{eqn:covarianceLayout}
    \Sigma = \begin{bmatrix}
    I_{n \times n} & 0 & 0 & \frac{\partial \tau}{\partial q}\\
    0& I_{n \times n} & 0 & \frac{\partial\tau}{\partial \dot{q}}\\
    0& 0 & I_{n \times n} & \frac{\partial \tau}{\partial \ddot{q}}\\
    \frac{\partial \tau}{\partial q} & \frac{\partial \tau}{\partial \dot{q}} & \frac{\partial \tau}{\partial \ddot{q}} & I_{n \times n}
    \end{bmatrix}
\end{equation}
Where $n$ is the number of degrees of freedom of the robot.

Using the fact the covariance should describe how those variables should change with respect to the other dimensions, the off-diagonal terms should be equal to the partial derivatives of those dimensions. In the context of this paper, we can omit the effect of the diagonal and off-diagonal terms on $q$, $\dot{q}$, $\ddot{q}$ as they should not change with respect to the inertial parameters. The important terms to consider are those on the off-diagonal terms involving the $\tau$.

Using \eqref{eqn:covarianceLayout} as an assumption of the covariance layout, the update to the covariance matrix is as follows:
\begin{equation}
    \mathbf{T} = \begin{bmatrix}
    I_{n \times n}& 0 & 0 & -\frac{\partial{\mathbf{Y}\left(q_{\mu},\dot{q}_{\mu},\ddot{q}_{\mu}\right)}}{\partial q} \Delta{\hat{\pi}}\\
    0& I_{n \times n} & 0 & -\frac{\partial{\mathbf{Y}\left(q_{\mu},\dot{q}_{\mu},\ddot{q}_{\mu}\right)}}{\partial \dot{q}} \Delta{\hat{\pi}}\\
    0& 0 & I_{n \times n} & -\frac{\partial{\mathbf{Y}\left(q_{\mu},\dot{q}_{\mu},\ddot{q}_{\mu}\right)}}{\partial \ddot{q}} \Delta{\hat{\pi}}\\
    0& 0 & 0 & I_{n \times n} 
    \end{bmatrix}^{T}
\end{equation}

The covariance matrix can be updated by this matrix by:
\begin{equation}
    \Sigma^{\prime} = \mathbf{T} \Sigma \mathbf{T}^T
\end{equation}

Using these equations with a trained GMM model you can maintain the consistency between the two Semi-Parametric components, and freely update both online and simultaneously with a minimal conflict between the components or invalidating the previously learned GMM model.

It is important to note at this point that the mean update is exact, however, the covariance update is only approximate in nature due to the Gaussian sub-models being a first-order approximation to the data. This can mean that if the Gaussian approximates an area where the inertial parameters have a large affect the approximation may deteriorate with large changes of these parameters.
\subsection{Implementation}
The implementation of this system was done using the OROCOS framework~\cite{Bruyninckx2003TheProject} which specializes in real-time execution. The dynamics matrices, regressors, and state derivatives of the regressors have been calculated using the Adaptive Robotics Dynamics Library (ARDL)\footnote{https://github.com/smithjoshua001/ARDL}, specializing in support for adaptive dynamics algorithms, using algorithms inspired by~\cite{Garofalo2013}. The velocity and acceleration of the joints were estimated through the use of a single-dimensional Kalman filter and PLL filter respectively~\cite{Wang2012VelocityVariance}.

\begin{figure}[h!]
  \centering
\begin{tikzpicture}[scale = 0.65,transform shape]

\node[block, name=trajectory] at(0,0) {Trajectory};

\node[block, name=slotineError, right = of trajectory, xshift=15mm] {Error Compute};
\node[block, name=CM, above = of slotineError] {Current State Model};
\node[branch, name=ISB,  above = of trajectory.-10] {};
\node[guide, name=inputstate, left = of ISB, xshift=-7.83mm] {};
\node[guide, name=ITB,  above = of trajectory.10] {};
\node[guide, name=inputtorque, left = of ITB, xshift=-8mm] {};

\node[block, name=P, right = of slotineError] {Parametric};
\node[name=TE,sum, right = of CM,scale=2] {};
\node[block,name=NP, right = of TE, xshift=-2mm] {Non-Parametric};

\node[guide, name=TSguide1, right = of NP] {};
\node[name=TS,sum,scale=2, right = of $(TSguide1|-P)!0.5!(P|-TSguide1)$] {};
\node[guide, name=out, right = of TS] {};
\node[branch, name=ISB1] at (CM.180-|ISB) {};
\node[guide, name=ISB2, above = of ISB1, yshift=-3mm] {};

\draw [connector]  (inputstate.0) --  node[below] {$q,\dot{q},\ddot{q}$}(ISB) -- ($(slotineError.175|-ISB)!0.2!(ISB)$)--  ($(slotineError.175|-ISB)!0.2!(ISB|-slotineError.175)$)|-(slotineError.175);

\draw [connector](inputtorque.center) -- node[above] {$\tau$} (ITB.center) -- ($(slotineError.185|-ITB.center)!0.3!(ITB.center)$) --  ($(slotineError.185|-ITB.center)!0.3!(ITB.center|-slotineError.185)$)|-(slotineError.185);
\draw[connector] (slotineError.5) -- node[above] {$\dot{q}_r$}($(P.180) +(slotineError.5)-(slotineError.0)$);
\draw[connector] (slotineError.355) -- node[below] {$\ddot{q}_r$}($(P.180) +(slotineError.355)-(slotineError.0)$);
\draw [connector] (P) -- (TS|-P)-|(TS);
\draw [connector] (NP) -- (TS|-NP)-|(TS);
\node[branch, name=torqueGuide0] at ($(slotineError.185|-ITB)!0.3!(ITB)$) {};
\draw [connector] (torqueGuide0.center) --(torqueGuide0.center-|TE) -- (TE) node[pos=0.8, left] {$-$};
\draw [connector] (ISB) -- (ISB|-CM.180) -- (CM.180);
\draw [connector] (CM) -- node[above]{$\tau_{c}$}(TE);
\draw [connector] (TE) -- node[pos=0.7,below] {$\tilde{\tau}$} (NP);
\draw [connector] (ISB1) -- (ISB1|-ISB2) -| ($(NP.170)!0.3!(TE|-NP.170)$) |- (NP.170);

\node[branch, name=TrB, yshift=-7mm] at (trajectory.0 -| slotineError.210) {};
\node[guide, name=TrB2, xshift=15mm] at (trajectory.0) {};
\draw[connector] (trajectory.0) -- node[pos=0.5,below] {$q_d, \dot{q}_d, \ddot{q}_d$} (TrB2.center) -- (TrB2.center|-TrB) -- (TrB) -- (slotineError.210);

\draw[connector] (TrB) -- (TrB-|NP.240) -- (NP.240);

\node[branch, name=STP] at ($(slotineError.175|-ISB)!0.2!(ISB)$) {};
\draw[connector] (STP) -- (STP-|P.140) -- (P.140);

\node[guide, name=TO, right = of TS] {};
\draw[connector] (TS.0) -- node[above] {$\tau_{\text{output}}$}(TO);

\draw[connector] (P.40) -- ($(P.40)!0.4!(P.40|-NP.200)$) -| node[xshift=-1mm,right] {$\Delta{\pi}$} (NP.200) --  (NP.200);
\end{tikzpicture}
\caption{System Diagram}\label{fig:systemDiagram}
\end{figure}
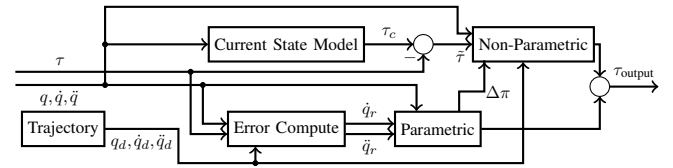

Figure \ref{fig:systemDiagram} shows the control system in the entirety with each component implemented as described in section \ref{sec:Methodology} with only the key input and outputs shown. It is key to note that the current model component is using the same model as the Parametric component and replicates the change in the inertial parameters.

\section{Experimental Validation}
To demonstrate the use of the consistency transformation we define two main experiments. The first will use a pre-trained baseline model \cite{Camoriano2016} and a pre-trained GMM model trained solely on artificial dynamics model data. The second shall train the proposed model with the transformation online on a real platform.

\subsection{Virtual Model Experiment}\label{subsec:VME}
To demonstrate the non-stationary effect, we shall use a virtual dynamics model of a Kuka LWR 4 manipulator, with simulated inertial parameters ($\pi$). This model will generate the torque data required to train both components and to provide the updated torque when the inertial parameters are changed. State data will be generated from seven pre-defined exciting trajectories appended to the end of each other. This state will contain $q$, $\dot{q}$, $\ddot{q}$, and $\tau$.

Initially, the Parametric component uses $0\pi$ for the inertial parameters, such that it produces zero torque. The baseline~\cite{Camoriano2016} and GMM model are then trained on the error as stated in \eqref{eqn:tauError} using every third data point. Effectively the models are trained to learn the entire robot dynamics.

The Parametric component is then updated to use $0.5\pi$ as inertial parameters. We then compare the output of the Parametric component plus the baseline/GMM model with the virtual dynamics model to get the normalized Mean Squared Error (nMSE) measure \eqref{eqn:nmse}. The consistency update is then applied to the GMM model and the outputs compared again.

\begin{equation}
  \text{nMSE}_{\bullet} = \frac{\frac{\sum (\bullet - \hat{\bullet})^2}{N}}{(\bullet_{\text{max}} - \bullet_{\text{min}})}\label{eqn:nmse}
\end{equation}
Where $\bullet_{\text{min}}$ and $\bullet_{\text{max}}$ are obtained as the maximum and minimum values of the desired position and velocity when $\bullet$ is $q$ or $\dot{q}$. When $\bullet$ is $\tau$ however, $\bullet_{\text{min}}$ and $\bullet_{\text{max}}$ are obtained as the maximum and minimum values of the recorded measured joint torque of the robot. 

The metrics that will then be analysed will be the initial baseline and GMM torque prediction error on the full dynamics, and the Semi-Parametric torque prediction error with and without the transformed GMM and the baseline model with the new inputs. The trajectories that will be used will be several different Modified Fourier trajectories~\cite{Park2006Fourier-basedRobots}, a cyclic trajectory that guarantees a particular starting state.







\subsection{Real Platform Experiment}\label{subsec:RPE}

To demonstrate the online simultaneous Semi-Parametric controller we aim to show three key features. First, the Non-Parametric component can learn the desired function, online, starting with an incorrect Parametric component. Second, applying the transform without updating the Non-Parametric component does not significantly increase the tracking or torque errors. Third, to show that both components can be learned simultaneously with the same effect.

To do this we use a Kuka~LWR~IV to execute five repeating trajectories for 10 minutes. The following phases are designed to show the initial errors due to the initial incorrect model, followed by the previously defined features.
\begin{enumerate}
    \item Starting from an incorrect Parametric component and an empty Non-Parametric component, the trajectory will be executed. The Non-Parametric component will be updated iteratively but not output to the control. The Parametric component does not update. (0-90 seconds)
    \item The Non-Parametric component will then continue to be updated iteratively and will output the learned torque error. The Parametric component does not update. (90-180 seconds)
    \item \label{enum:transformationStep} The Parametric component will then be updated iteratively with the Non-Parametric component being transformed accordingly but not updated. (180-360 seconds) 
    \item The Parametric component will continue to be updated. The Non-Parametric component will both be updated iteratively and transformed continuously. (360-600 seconds)
\end{enumerate}

The phases will also be modified to show the effects of not applying the consistency transform. In particular, phase \ref{enum:transformationStep} will become identical to phase 4.

With the real robot, due to the lack of ground truth of the dynamics at the desired state, ($q_d, \dot{q}_d, \ddot{q}_d$), we cannot directly use the torque as an performance metric. 
We use other metrics as indirect indicators of performance. In particular, we shall look at the state errors for position and velocity, which provide the feedback signal for the controller, and the error between the measured torque and the estimate of the current torque from the Semi-Parametric model at the robot's current state.

The exemplar trajectory, shown in \cref{fig:snaps}, will be shown in a graph with and without the transform. The main evidence will be provided as nMSE of the performance metrics, calculated from the concatenated data from each phase of each trajectory. The desired outcome of this experiment should show minimal change to the normalized mean squared error (nMSE) or a reduced nMSE.

The parameters for the Parametric component are as follows: $\mathbf{R} = I_{n \times n}$, $\mathbf{P}$ is obtained through equations (16) and (17) in \cite{Slotine1989}, with $\lambda_0 = 1.5$ and $k_0 = 0.1$. The error gains $\mathbold{\Lambda} =$ diag($[20,20,20,20,10,10,10]$) and $\mathbf{K}_D = $ diag($[5,5,5,5,2,2,2]$) from (6) and (8) in \cite{Slotine1989}. The initial $\Sigma$ for the IGMM algorithm was obtained through a subset of pre-collected data from the trajectory shown in \cref{fig:snaps}.

\section{Results}
\begin{figure*}
  \centering

\begingroup
  \makeatletter
  \providecommand\color[2][]{%
    \GenericError{(gnuplot) \space\space\space\@spaces}{%
      Package color not loaded in conjunction with
      terminal option `colourtext'%
    }{See the gnuplot documentation for explanation.%
    }{Either use 'blacktext' in gnuplot or load the package
      color.sty in LaTeX.}%
    \renewcommand\color[2][]{}%
  }%
  \providecommand\includegraphics[2][]{%
    \GenericError{(gnuplot) \space\space\space\@spaces}{%
      Package graphicx or graphics not loaded%
    }{See the gnuplot documentation for explanation.%
    }{The gnuplot epslatex terminal needs graphicx.sty or graphics.sty.}%
    \renewcommand\includegraphics[2][]{}%
  }%
  \providecommand\rotatebox[2]{#2}%
  \@ifundefined{ifGPcolor}{%
    \newif\ifGPcolor
    \GPcolortrue
  }{}%
  \@ifundefined{ifGPblacktext}{%
    \newif\ifGPblacktext
    \GPblacktexttrue
  }{}%
  \let\gplgaddtomacro\g@addto@macro
  \gdef\gplbacktext{}%
  \gdef\gplfronttext{}%
  \makeatother
  \ifGPblacktext
    \def\colorrgb#1{}%
    \def\colorgray#1{}%
  \else
    \ifGPcolor
      \def\colorrgb#1{\color[rgb]{#1}}%
      \def\colorgray#1{\color[gray]{#1}}%
      \expandafter\def\csname LTw\endcsname{\color{white}}%
      \expandafter\def\csname LTb\endcsname{\color{black}}%
      \expandafter\def\csname LTa\endcsname{\color{black}}%
      \expandafter\def\csname LT0\endcsname{\color[rgb]{1,0,0}}%
      \expandafter\def\csname LT1\endcsname{\color[rgb]{0,1,0}}%
      \expandafter\def\csname LT2\endcsname{\color[rgb]{0,0,1}}%
      \expandafter\def\csname LT3\endcsname{\color[rgb]{1,0,1}}%
      \expandafter\def\csname LT4\endcsname{\color[rgb]{0,1,1}}%
      \expandafter\def\csname LT5\endcsname{\color[rgb]{1,1,0}}%
      \expandafter\def\csname LT6\endcsname{\color[rgb]{0,0,0}}%
      \expandafter\def\csname LT7\endcsname{\color[rgb]{1,0.3,0}}%
      \expandafter\def\csname LT8\endcsname{\color[rgb]{0.5,0.5,0.5}}%
    \else
      \def\colorrgb#1{\color{black}}%
      \def\colorgray#1{\color[gray]{#1}}%
      \expandafter\def\csname LTw\endcsname{\color{white}}%
      \expandafter\def\csname LTb\endcsname{\color{black}}%
      \expandafter\def\csname LTa\endcsname{\color{black}}%
      \expandafter\def\csname LT0\endcsname{\color{black}}%
      \expandafter\def\csname LT1\endcsname{\color{black}}%
      \expandafter\def\csname LT2\endcsname{\color{black}}%
      \expandafter\def\csname LT3\endcsname{\color{black}}%
      \expandafter\def\csname LT4\endcsname{\color{black}}%
      \expandafter\def\csname LT5\endcsname{\color{black}}%
      \expandafter\def\csname LT6\endcsname{\color{black}}%
      \expandafter\def\csname LT7\endcsname{\color{black}}%
      \expandafter\def\csname LT8\endcsname{\color{black}}%
    \fi
  \fi
    \setlength{\unitlength}{0.0500bp}%
    \ifx\gptboxheight\undefined%
      \newlength{\gptboxheight}%
      \newlength{\gptboxwidth}%
      \newsavebox{\gptboxtext}%
    \fi%
    \setlength{\fboxrule}{0.5pt}%
    \setlength{\fboxsep}{1pt}%
\begin{picture}(9060.00,2180.00)%
    \gplgaddtomacro\gplbacktext{%
      \csname LTb\endcsname
      \put(951,464){\makebox(0,0)[r]{\strut{}1e-06}}%
      \csname LTb\endcsname
      \put(951,632){\makebox(0,0)[r]{\strut{}1e-05}}%
      \csname LTb\endcsname
      \put(951,800){\makebox(0,0)[r]{\strut{}0.0001}}%
      \csname LTb\endcsname
      \put(951,968){\makebox(0,0)[r]{\strut{}0.001}}%
      \csname LTb\endcsname
      \put(951,1136){\makebox(0,0)[r]{\strut{}0.01}}%
      \csname LTb\endcsname
      \put(951,1303){\makebox(0,0)[r]{\strut{}0.1}}%
      \csname LTb\endcsname
      \put(951,1471){\makebox(0,0)[r]{\strut{}1}}%
      \csname LTb\endcsname
      \put(951,1639){\makebox(0,0)[r]{\strut{}10}}%
      \csname LTb\endcsname
      \put(951,1807){\makebox(0,0)[r]{\strut{}100}}%
      \csname LTb\endcsname
      \put(1053,278){\makebox(0,0){\strut{}}}%
      \csname LTb\endcsname
      \put(2016,278){\makebox(0,0){\strut{}}}%
      \csname LTb\endcsname
      \put(2978,278){\makebox(0,0){\strut{}}}%
      \csname LTb\endcsname
      \put(3941,278){\makebox(0,0){\strut{}}}%
      \csname LTb\endcsname
      \put(4903,278){\makebox(0,0){\strut{}}}%
      \csname LTb\endcsname
      \put(5866,278){\makebox(0,0){\strut{}}}%
      \csname LTb\endcsname
      \put(6828,278){\makebox(0,0){\strut{}}}%
      \csname LTb\endcsname
      \put(7791,278){\makebox(0,0){\strut{}}}%
      \csname LTb\endcsname
      \put(8753,278){\makebox(0,0){\strut{}}}%
      \csname LTb\endcsname
      \put(2016,1993){\makebox(0,0){\strut{}Joint 1}}%
      \csname LTb\endcsname
      \put(2978,1993){\makebox(0,0){\strut{}Joint 2}}%
      \csname LTb\endcsname
      \put(3941,1993){\makebox(0,0){\strut{}Joint 3}}%
      \csname LTb\endcsname
      \put(4903,1993){\makebox(0,0){\strut{}Joint 4}}%
      \csname LTb\endcsname
      \put(5866,1993){\makebox(0,0){\strut{}Joint 5}}%
      \csname LTb\endcsname
      \put(6828,1993){\makebox(0,0){\strut{}Joint 6}}%
      \csname LTb\endcsname
      \put(7791,1993){\makebox(0,0){\strut{}Joint 7}}%
    }%
    \gplgaddtomacro\gplfronttext{%
      \csname LTb\endcsname
      \put(306,1135){\rotatebox{-270}{\makebox(0,0){\strut{}nMSE$_{\tau}$}}}%
      \csname LTb\endcsname
      \put(1433,282){\makebox(0,0)[l]{\strut{}\scriptsize \cite{Camoriano2016}  ($0\pi$)}}%
      \csname LTb\endcsname
      \put(1433,143){\makebox(0,0)[l]{\strut{}\scriptsize \cite{Camoriano2016}  ($0.5\pi$)}}%
      \csname LTb\endcsname
      \put(4057,282){\makebox(0,0)[l]{\strut{}\scriptsize Proposed ($0\pi$)}}%
      \csname LTb\endcsname
      \put(4057,143){\makebox(0,0)[l]{\strut{}\scriptsize Proposed  ($0.5\pi$) without transform}}%
      \csname LTb\endcsname
      \put(6681,282){\makebox(0,0)[l]{\strut{}\scriptsize Proposed  ($0.5\pi$) with transform}}%
    }%
    \gplbacktext
    \put(0,0){\includegraphics{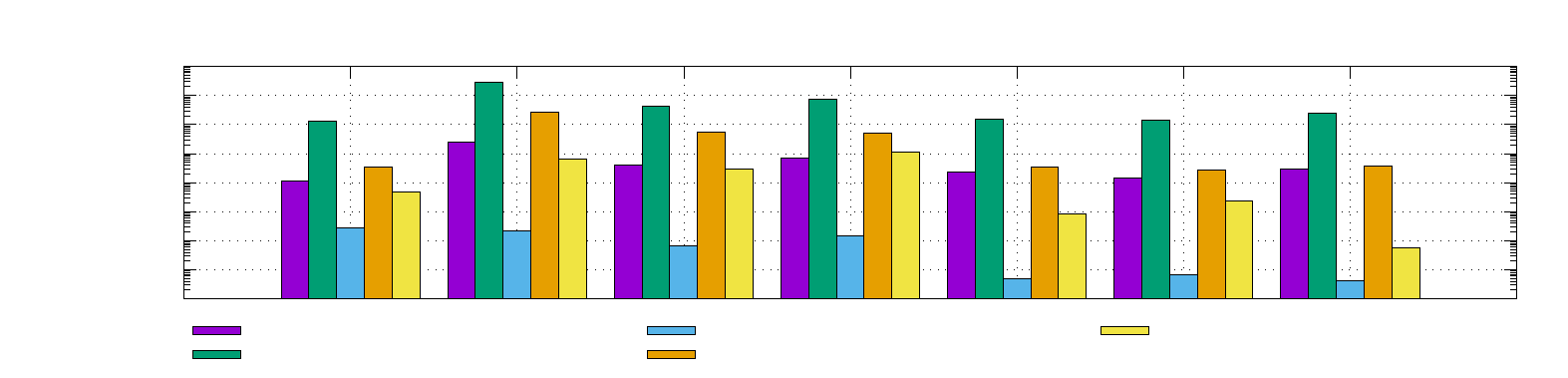}}%
    \gplfronttext
  \end{picture}%
\endgroup
\vspace*{-1.0\baselineskip}
\caption{Virtual Model Adaptation - The random feature RRLS model \cite{Camoriano2016}, shows the non-stationary issue, as evident in the large increase in nMSE in all joints between the $0\pi$ and $0.5\pi$ models. The proposed GMM model demonstrates the same effect between the two models without consistency transform, however, when the consistency transform is applied the error is reduced. This is a strong indication that the model is kept more consistent with the update in inertial parameters.}\label{fig:VVMA} 
\end{figure*}
\subsection{Transformation Validation}\label{ssec:TV}

The results from the simulated experiment are shown in \cref{fig:VVMA}. The chart shows three main results; the errors for the baseline model \cite{Camoriano2016}, with 400 random features, and the GMM model initially trained on the full torque, the recall errors for both models when the inertial parameter update is applied, and the recall error for the GMM model transformed with the consistency transform.

The results show that the baseline method \cite{Camoriano2016} and the GMM model without the consistency transform demonstrate the non-stationary effect increasing the nMSE in all joints.
The GMM model with the consistency transform does show an increase in error from the original model. This is reduced compared to the baseline and GMM without the transform models. This shows that the transformation, whilst not perfect due to the approximate nature of the transform, adjusts the GMM to the change in parameters allowing the old data to remain more consistent. 
\begin{figure*}
    \vspace*{.5\baselineskip}
    \captionsetup[subfigure]{aboveskip=-1\baselineskip,belowskip=0.4\baselineskip}
    \centering
\begin{subfigure}{\textwidth}
    \centering

\begingroup
  \makeatletter
  \providecommand\color[2][]{%
    \GenericError{(gnuplot) \space\space\space\@spaces}{%
      Package color not loaded in conjunction with
      terminal option `colourtext'%
    }{See the gnuplot documentation for explanation.%
    }{Either use 'blacktext' in gnuplot or load the package
      color.sty in LaTeX.}%
    \renewcommand\color[2][]{}%
  }%
  \providecommand\includegraphics[2][]{%
    \GenericError{(gnuplot) \space\space\space\@spaces}{%
      Package graphicx or graphics not loaded%
    }{See the gnuplot documentation for explanation.%
    }{The gnuplot epslatex terminal needs graphicx.sty or graphics.sty.}%
    \renewcommand\includegraphics[2][]{}%
  }%
  \providecommand\rotatebox[2]{#2}%
  \@ifundefined{ifGPcolor}{%
    \newif\ifGPcolor
    \GPcolortrue
  }{}%
  \@ifundefined{ifGPblacktext}{%
    \newif\ifGPblacktext
    \GPblacktexttrue
  }{}%
  \let\gplgaddtomacro\g@addto@macro
  \gdef\gplbacktext{}%
  \gdef\gplfronttext{}%
  \makeatother
  \ifGPblacktext
    \def\colorrgb#1{}%
    \def\colorgray#1{}%
  \else
    \ifGPcolor
      \def\colorrgb#1{\color[rgb]{#1}}%
      \def\colorgray#1{\color[gray]{#1}}%
      \expandafter\def\csname LTw\endcsname{\color{white}}%
      \expandafter\def\csname LTb\endcsname{\color{black}}%
      \expandafter\def\csname LTa\endcsname{\color{black}}%
      \expandafter\def\csname LT0\endcsname{\color[rgb]{1,0,0}}%
      \expandafter\def\csname LT1\endcsname{\color[rgb]{0,1,0}}%
      \expandafter\def\csname LT2\endcsname{\color[rgb]{0,0,1}}%
      \expandafter\def\csname LT3\endcsname{\color[rgb]{1,0,1}}%
      \expandafter\def\csname LT4\endcsname{\color[rgb]{0,1,1}}%
      \expandafter\def\csname LT5\endcsname{\color[rgb]{1,1,0}}%
      \expandafter\def\csname LT6\endcsname{\color[rgb]{0,0,0}}%
      \expandafter\def\csname LT7\endcsname{\color[rgb]{1,0.3,0}}%
      \expandafter\def\csname LT8\endcsname{\color[rgb]{0.5,0.5,0.5}}%
    \else
      \def\colorrgb#1{\color{black}}%
      \def\colorgray#1{\color[gray]{#1}}%
      \expandafter\def\csname LTw\endcsname{\color{white}}%
      \expandafter\def\csname LTb\endcsname{\color{black}}%
      \expandafter\def\csname LTa\endcsname{\color{black}}%
      \expandafter\def\csname LT0\endcsname{\color{black}}%
      \expandafter\def\csname LT1\endcsname{\color{black}}%
      \expandafter\def\csname LT2\endcsname{\color{black}}%
      \expandafter\def\csname LT3\endcsname{\color{black}}%
      \expandafter\def\csname LT4\endcsname{\color{black}}%
      \expandafter\def\csname LT5\endcsname{\color{black}}%
      \expandafter\def\csname LT6\endcsname{\color{black}}%
      \expandafter\def\csname LT7\endcsname{\color{black}}%
      \expandafter\def\csname LT8\endcsname{\color{black}}%
    \fi
  \fi
    \setlength{\unitlength}{0.0500bp}%
    \ifx\gptboxheight\undefined%
      \newlength{\gptboxheight}%
      \newlength{\gptboxwidth}%
      \newsavebox{\gptboxtext}%
    \fi%
    \setlength{\fboxrule}{0.5pt}%
    \setlength{\fboxsep}{1pt}%
\begin{picture}(10080.00,2720.00)%
    \gplgaddtomacro\gplbacktext{%
      \csname LTb\endcsname
      \put(951,837){\makebox(0,0)[r]{\strut{}1e-06}}%
      \csname LTb\endcsname
      \put(951,1079){\makebox(0,0)[r]{\strut{}1e-05}}%
      \csname LTb\endcsname
      \put(951,1322){\makebox(0,0)[r]{\strut{}0.0001}}%
      \csname LTb\endcsname
      \put(951,1564){\makebox(0,0)[r]{\strut{}0.001}}%
      \csname LTb\endcsname
      \put(951,1806){\makebox(0,0)[r]{\strut{}0.01}}%
      \csname LTb\endcsname
      \put(951,2048){\makebox(0,0)[r]{\strut{}0.1}}%
      \csname LTb\endcsname
      \put(951,2291){\makebox(0,0)[r]{\strut{}1}}%
      \csname LTb\endcsname
      \put(951,2533){\makebox(0,0)[r]{\strut{}10}}%
      \csname LTb\endcsname
      \put(1364,651){\makebox(0,0){\strut{}\scriptsize $q$}}%
      \csname LTb\endcsname
      \put(1676,651){\makebox(0,0){\strut{}\scriptsize $\dot{q}$}}%
      \csname LTb\endcsname
      \put(1987,651){\makebox(0,0){\strut{}\scriptsize $\tau$}}%
      \csname LTb\endcsname
      \put(2610,651){\makebox(0,0){\strut{}\scriptsize $q$}}%
      \csname LTb\endcsname
      \put(2922,651){\makebox(0,0){\strut{}\scriptsize $\dot{q}$}}%
      \csname LTb\endcsname
      \put(3233,651){\makebox(0,0){\strut{}\scriptsize $\tau$}}%
      \csname LTb\endcsname
      \put(3856,651){\makebox(0,0){\strut{}\scriptsize $q$}}%
      \csname LTb\endcsname
      \put(4167,651){\makebox(0,0){\strut{}\scriptsize $\dot{q}$}}%
      \csname LTb\endcsname
      \put(4479,651){\makebox(0,0){\strut{}\scriptsize $\tau$}}%
      \csname LTb\endcsname
      \put(5102,651){\makebox(0,0){\strut{}\scriptsize $q$}}%
      \csname LTb\endcsname
      \put(5413,651){\makebox(0,0){\strut{}\scriptsize $\dot{q}$}}%
      \csname LTb\endcsname
      \put(5724,651){\makebox(0,0){\strut{}\scriptsize $\tau$}}%
      \csname LTb\endcsname
      \put(6347,651){\makebox(0,0){\strut{}\scriptsize $q$}}%
      \csname LTb\endcsname
      \put(6659,651){\makebox(0,0){\strut{}\scriptsize $\dot{q}$}}%
      \csname LTb\endcsname
      \put(6970,651){\makebox(0,0){\strut{}\scriptsize $\tau$}}%
      \csname LTb\endcsname
      \put(7593,651){\makebox(0,0){\strut{}\scriptsize $q$}}%
      \csname LTb\endcsname
      \put(7904,651){\makebox(0,0){\strut{}\scriptsize $\dot{q}$}}%
      \csname LTb\endcsname
      \put(8216,651){\makebox(0,0){\strut{}\scriptsize $\tau$}}%
      \csname LTb\endcsname
      \put(8839,651){\makebox(0,0){\strut{}\scriptsize $q$}}%
      \csname LTb\endcsname
      \put(9150,651){\makebox(0,0){\strut{}\scriptsize $\dot{q}$}}%
      \csname LTb\endcsname
      \put(9462,651){\makebox(0,0){\strut{}\scriptsize $\tau$}}%
    }%
    \gplgaddtomacro\gplfronttext{%
      \csname LTb\endcsname
      \put(306,1685){\rotatebox{-270}{\makebox(0,0){\strut{}nMSE}}}%
      \csname LTb\endcsname
      \put(3123,353){\makebox(0,0)[r]{\strut{}\scriptsize Phase 1}}%
      \csname LTb\endcsname
      \put(4727,353){\makebox(0,0)[r]{\strut{}\scriptsize Phase 2}}%
      \csname LTb\endcsname
      \put(6331,353){\makebox(0,0)[r]{\strut{}\scriptsize Phase 3}}%
      \csname LTb\endcsname
      \put(7935,353){\makebox(0,0)[r]{\strut{}\scriptsize Phase 4}}%
      \csname LTb\endcsname
      \put(9150,2594){\makebox(0,0){\strut{}Joint 7}}%
      \csname LTb\endcsname
      \put(7904,2594){\makebox(0,0){\strut{}Joint 6}}%
      \csname LTb\endcsname
      \put(6659,2594){\makebox(0,0){\strut{}Joint 5}}%
      \csname LTb\endcsname
      \put(5413,2594){\makebox(0,0){\strut{}Joint 4}}%
      \csname LTb\endcsname
      \put(4167,2594){\makebox(0,0){\strut{}Joint 3}}%
      \csname LTb\endcsname
      \put(2922,2594){\makebox(0,0){\strut{}Joint 2}}%
      \csname LTb\endcsname
      \put(1676,2594){\makebox(0,0){\strut{}Joint 1}}%
    }%
    \gplbacktext
    \put(0,0){\includegraphics{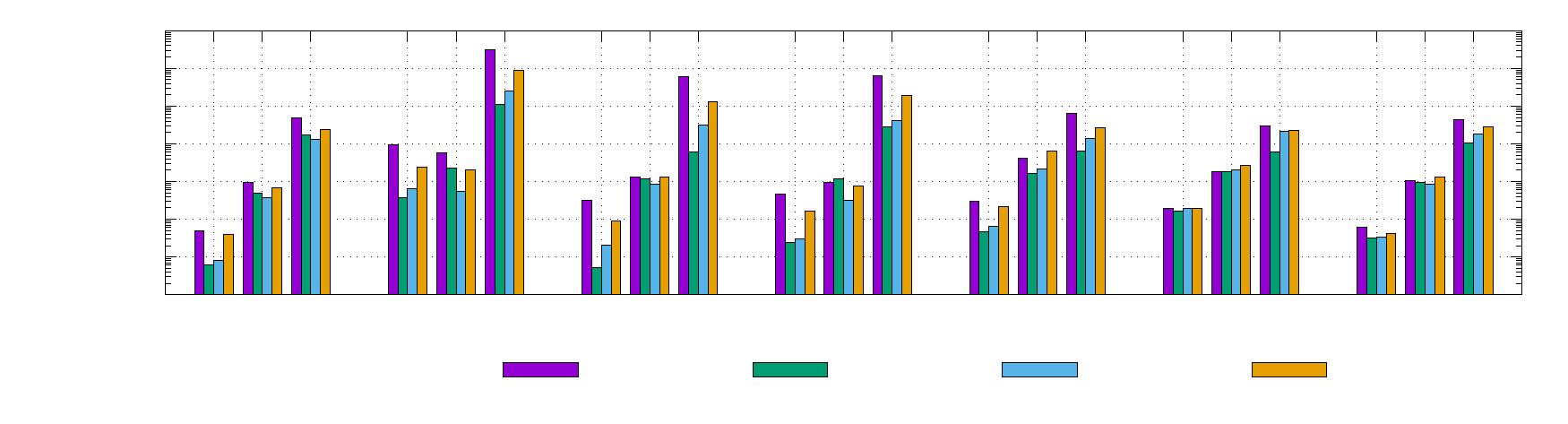}}%
    \gplfronttext
  \end{picture}%
\endgroup
    \vspace*{-0\baselineskip}
    \caption{No Transform (logscale) - With no transform the nMSE appears to have a trend where Phase 2 learns the dynamics decreasing the nMSE error. Onwards though through Phase 3 and 4 the nMSE tends to increase in most joints with the final nMSE in all joints over all metrics being larger than in Phase 2. This indicates the two components conflicting.}\label{fig:RRNT}
\end{subfigure}
\begin{subfigure}{\textwidth}
    \centering


\begingroup
  \makeatletter
  \providecommand\color[2][]{%
    \GenericError{(gnuplot) \space\space\space\@spaces}{%
      Package color not loaded in conjunction with
      terminal option `colourtext'%
    }{See the gnuplot documentation for explanation.%
    }{Either use 'blacktext' in gnuplot or load the package
      color.sty in LaTeX.}%
    \renewcommand\color[2][]{}%
  }%
  \providecommand\includegraphics[2][]{%
    \GenericError{(gnuplot) \space\space\space\@spaces}{%
      Package graphicx or graphics not loaded%
    }{See the gnuplot documentation for explanation.%
    }{The gnuplot epslatex terminal needs graphicx.sty or graphics.sty.}%
    \renewcommand\includegraphics[2][]{}%
  }%
  \providecommand\rotatebox[2]{#2}%
  \@ifundefined{ifGPcolor}{%
    \newif\ifGPcolor
    \GPcolortrue
  }{}%
  \@ifundefined{ifGPblacktext}{%
    \newif\ifGPblacktext
    \GPblacktexttrue
  }{}%
  \let\gplgaddtomacro\g@addto@macro
  \gdef\gplbacktext{}%
  \gdef\gplfronttext{}%
  \makeatother
  \ifGPblacktext
    \def\colorrgb#1{}%
    \def\colorgray#1{}%
  \else
    \ifGPcolor
      \def\colorrgb#1{\color[rgb]{#1}}%
      \def\colorgray#1{\color[gray]{#1}}%
      \expandafter\def\csname LTw\endcsname{\color{white}}%
      \expandafter\def\csname LTb\endcsname{\color{black}}%
      \expandafter\def\csname LTa\endcsname{\color{black}}%
      \expandafter\def\csname LT0\endcsname{\color[rgb]{1,0,0}}%
      \expandafter\def\csname LT1\endcsname{\color[rgb]{0,1,0}}%
      \expandafter\def\csname LT2\endcsname{\color[rgb]{0,0,1}}%
      \expandafter\def\csname LT3\endcsname{\color[rgb]{1,0,1}}%
      \expandafter\def\csname LT4\endcsname{\color[rgb]{0,1,1}}%
      \expandafter\def\csname LT5\endcsname{\color[rgb]{1,1,0}}%
      \expandafter\def\csname LT6\endcsname{\color[rgb]{0,0,0}}%
      \expandafter\def\csname LT7\endcsname{\color[rgb]{1,0.3,0}}%
      \expandafter\def\csname LT8\endcsname{\color[rgb]{0.5,0.5,0.5}}%
    \else
      \def\colorrgb#1{\color{black}}%
      \def\colorgray#1{\color[gray]{#1}}%
      \expandafter\def\csname LTw\endcsname{\color{white}}%
      \expandafter\def\csname LTb\endcsname{\color{black}}%
      \expandafter\def\csname LTa\endcsname{\color{black}}%
      \expandafter\def\csname LT0\endcsname{\color{black}}%
      \expandafter\def\csname LT1\endcsname{\color{black}}%
      \expandafter\def\csname LT2\endcsname{\color{black}}%
      \expandafter\def\csname LT3\endcsname{\color{black}}%
      \expandafter\def\csname LT4\endcsname{\color{black}}%
      \expandafter\def\csname LT5\endcsname{\color{black}}%
      \expandafter\def\csname LT6\endcsname{\color{black}}%
      \expandafter\def\csname LT7\endcsname{\color{black}}%
      \expandafter\def\csname LT8\endcsname{\color{black}}%
    \fi
  \fi
    \setlength{\unitlength}{0.0500bp}%
    \ifx\gptboxheight\undefined%
      \newlength{\gptboxheight}%
      \newlength{\gptboxwidth}%
      \newsavebox{\gptboxtext}%
    \fi%
    \setlength{\fboxrule}{0.5pt}%
    \setlength{\fboxsep}{1pt}%
\begin{picture}(10080.00,2720.00)%
    \gplgaddtomacro\gplbacktext{%
      \csname LTb\endcsname
      \put(951,837){\makebox(0,0)[r]{\strut{}1e-06}}%
      \csname LTb\endcsname
      \put(951,1079){\makebox(0,0)[r]{\strut{}1e-05}}%
      \csname LTb\endcsname
      \put(951,1322){\makebox(0,0)[r]{\strut{}0.0001}}%
      \csname LTb\endcsname
      \put(951,1564){\makebox(0,0)[r]{\strut{}0.001}}%
      \csname LTb\endcsname
      \put(951,1806){\makebox(0,0)[r]{\strut{}0.01}}%
      \csname LTb\endcsname
      \put(951,2048){\makebox(0,0)[r]{\strut{}0.1}}%
      \csname LTb\endcsname
      \put(951,2291){\makebox(0,0)[r]{\strut{}1}}%
      \csname LTb\endcsname
      \put(951,2533){\makebox(0,0)[r]{\strut{}10}}%
      \csname LTb\endcsname
      \put(1364,651){\makebox(0,0){\strut{}\scriptsize $q$}}%
      \csname LTb\endcsname
      \put(1676,651){\makebox(0,0){\strut{}\scriptsize $\dot{q}$}}%
      \csname LTb\endcsname
      \put(1987,651){\makebox(0,0){\strut{}\scriptsize $\tau$}}%
      \csname LTb\endcsname
      \put(2610,651){\makebox(0,0){\strut{}\scriptsize $q$}}%
      \csname LTb\endcsname
      \put(2922,651){\makebox(0,0){\strut{}\scriptsize $\dot{q}$}}%
      \csname LTb\endcsname
      \put(3233,651){\makebox(0,0){\strut{}\scriptsize $\tau$}}%
      \csname LTb\endcsname
      \put(3856,651){\makebox(0,0){\strut{}\scriptsize $q$}}%
      \csname LTb\endcsname
      \put(4167,651){\makebox(0,0){\strut{}\scriptsize $\dot{q}$}}%
      \csname LTb\endcsname
      \put(4479,651){\makebox(0,0){\strut{}\scriptsize $\tau$}}%
      \csname LTb\endcsname
      \put(5102,651){\makebox(0,0){\strut{}\scriptsize $q$}}%
      \csname LTb\endcsname
      \put(5413,651){\makebox(0,0){\strut{}\scriptsize $\dot{q}$}}%
      \csname LTb\endcsname
      \put(5724,651){\makebox(0,0){\strut{}\scriptsize $\tau$}}%
      \csname LTb\endcsname
      \put(6347,651){\makebox(0,0){\strut{}\scriptsize $q$}}%
      \csname LTb\endcsname
      \put(6659,651){\makebox(0,0){\strut{}\scriptsize $\dot{q}$}}%
      \csname LTb\endcsname
      \put(6970,651){\makebox(0,0){\strut{}\scriptsize $\tau$}}%
      \csname LTb\endcsname
      \put(7593,651){\makebox(0,0){\strut{}\scriptsize $q$}}%
      \csname LTb\endcsname
      \put(7904,651){\makebox(0,0){\strut{}\scriptsize $\dot{q}$}}%
      \csname LTb\endcsname
      \put(8216,651){\makebox(0,0){\strut{}\scriptsize $\tau$}}%
      \csname LTb\endcsname
      \put(8839,651){\makebox(0,0){\strut{}\scriptsize $q$}}%
      \csname LTb\endcsname
      \put(9150,651){\makebox(0,0){\strut{}\scriptsize $\dot{q}$}}%
      \csname LTb\endcsname
      \put(9462,651){\makebox(0,0){\strut{}\scriptsize $\tau$}}%
    }%
    \gplgaddtomacro\gplfronttext{%
      \csname LTb\endcsname
      \put(306,1685){\rotatebox{-270}{\makebox(0,0){\strut{}nMSE}}}%
      \csname LTb\endcsname
      \put(3123,353){\makebox(0,0)[r]{\strut{}\scriptsize Phase 1}}%
      \csname LTb\endcsname
      \put(4727,353){\makebox(0,0)[r]{\strut{}\scriptsize Phase 2}}%
      \csname LTb\endcsname
      \put(6331,353){\makebox(0,0)[r]{\strut{}\scriptsize Phase 3}}%
      \csname LTb\endcsname
      \put(7935,353){\makebox(0,0)[r]{\strut{}\scriptsize Phase 4}}%
      \csname LTb\endcsname
      \put(9150,2594){\makebox(0,0){\strut{}Joint 7}}%
      \csname LTb\endcsname
      \put(7904,2594){\makebox(0,0){\strut{}Joint 6}}%
      \csname LTb\endcsname
      \put(6659,2594){\makebox(0,0){\strut{}Joint 5}}%
      \csname LTb\endcsname
      \put(5413,2594){\makebox(0,0){\strut{}Joint 4}}%
      \csname LTb\endcsname
      \put(4167,2594){\makebox(0,0){\strut{}Joint 3}}%
      \csname LTb\endcsname
      \put(2922,2594){\makebox(0,0){\strut{}Joint 2}}%
      \csname LTb\endcsname
      \put(1676,2594){\makebox(0,0){\strut{}Joint 1}}%
    }%
    \gplbacktext
    \put(0,0){\includegraphics{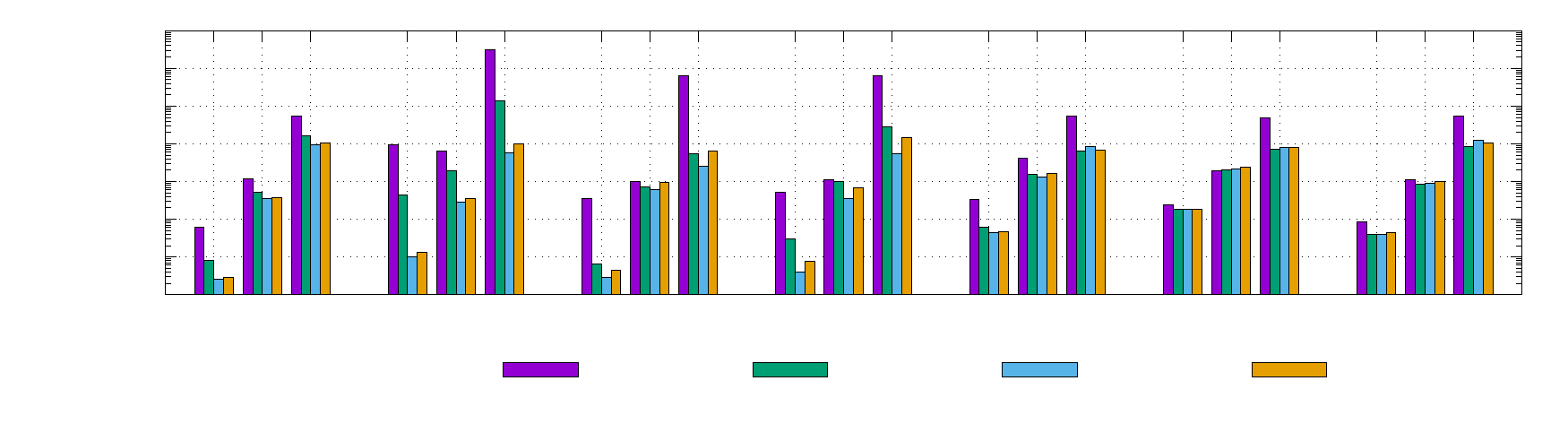}}%
    \gplfronttext
  \end{picture}%
\endgroup
    \vspace*{-0\baselineskip}
    \caption{Consistency Transform (logscale) - The metrics show that the nMSE reduces from phase 2. After phase two the nMSE does not change by a large margin, especially when comparing both Phases 3 and 4 against Phase 2. This is indicative that the two components are being kept consistent with each other, maintaining better accuracy overall.}\label{fig:RRCT}
\end{subfigure}
\vspace*{-0.9\baselineskip}
\caption{nMSE of metrics ($q$ (\si{\radian}), $\dot{q}$ (\si{\radian\per\second}), $\tau$ (\si{\newton\metre}))}\label{fig:LRMSE}
\end{figure*}
\subsection{Real Platform Experiment}\label{ssec:RPV}
\Cref{fig:LRMSE} displays the nMSE results, along a log scale, when running without and with the consistency transform respectively.

In \cref{fig:RRNT} we can see that during Phase 1, where the Non-Parametric component does not output, the errors are at the maximum they can be with respect to the high feedback gains. The torque nMSE in this phase is particularly high. During Phase 2 we can notice a drop in most of the metrics. The torque nMSE in Phase 2 again is quite high, especially in joint 2. Phases 3 and 4, which are equivalent for the no transformation scenario, show that nMSE increases almost back to the original nMSE from Phase 1. 

From \cref{fig:RRCT}, Phase 1 and 2 are very similar in scale to \cref{fig:RRNT}. Phases 3 and 4 we can see that the nMSE metrics are of a similar or less order than in Phase 2. There is a small increase that can be noticed in some metrics from Phase 3 to 4, however, these are relatively small and are most likely the cause of the GMM having local sub-models that sub-optimally represent that space in the target function or due to numerical instability on the covariance. The lack of improvement in the nMSE measures does not mean that the models are inconsistent in this case as it can be that the components, given their initial parameters, have reached a minima.

\begin{figure*}
    \captionsetup[subfigure]{aboveskip=-0.8\baselineskip,belowskip=0.2\baselineskip}
    \vspace*{\floatsep}
    \begin{subfigure}{\textwidth}
    \centering
    
\begingroup
  \makeatletter
  \providecommand\color[2][]{%
    \GenericError{(gnuplot) \space\space\space\@spaces}{%
      Package color not loaded in conjunction with
      terminal option `colourtext'%
    }{See the gnuplot documentation for explanation.%
    }{Either use 'blacktext' in gnuplot or load the package
      color.sty in LaTeX.}%
    \renewcommand\color[2][]{}%
  }%
  \providecommand\includegraphics[2][]{%
    \GenericError{(gnuplot) \space\space\space\@spaces}{%
      Package graphicx or graphics not loaded%
    }{See the gnuplot documentation for explanation.%
    }{The gnuplot epslatex terminal needs graphicx.sty or graphics.sty.}%
    \renewcommand\includegraphics[2][]{}%
  }%
  \providecommand\rotatebox[2]{#2}%
  \@ifundefined{ifGPcolor}{%
    \newif\ifGPcolor
    \GPcolortrue
  }{}%
  \@ifundefined{ifGPblacktext}{%
    \newif\ifGPblacktext
    \GPblacktexttrue
  }{}%
  \let\gplgaddtomacro\g@addto@macro
  \gdef\gplbacktext{}%
  \gdef\gplfronttext{}%
  \makeatother
  \ifGPblacktext
    \def\colorrgb#1{}%
    \def\colorgray#1{}%
  \else
    \ifGPcolor
      \def\colorrgb#1{\color[rgb]{#1}}%
      \def\colorgray#1{\color[gray]{#1}}%
      \expandafter\def\csname LTw\endcsname{\color{white}}%
      \expandafter\def\csname LTb\endcsname{\color{black}}%
      \expandafter\def\csname LTa\endcsname{\color{black}}%
      \expandafter\def\csname LT0\endcsname{\color[rgb]{1,0,0}}%
      \expandafter\def\csname LT1\endcsname{\color[rgb]{0,1,0}}%
      \expandafter\def\csname LT2\endcsname{\color[rgb]{0,0,1}}%
      \expandafter\def\csname LT3\endcsname{\color[rgb]{1,0,1}}%
      \expandafter\def\csname LT4\endcsname{\color[rgb]{0,1,1}}%
      \expandafter\def\csname LT5\endcsname{\color[rgb]{1,1,0}}%
      \expandafter\def\csname LT6\endcsname{\color[rgb]{0,0,0}}%
      \expandafter\def\csname LT7\endcsname{\color[rgb]{1,0.3,0}}%
      \expandafter\def\csname LT8\endcsname{\color[rgb]{0.5,0.5,0.5}}%
    \else
      \def\colorrgb#1{\color{black}}%
      \def\colorgray#1{\color[gray]{#1}}%
      \expandafter\def\csname LTw\endcsname{\color{white}}%
      \expandafter\def\csname LTb\endcsname{\color{black}}%
      \expandafter\def\csname LTa\endcsname{\color{black}}%
      \expandafter\def\csname LT0\endcsname{\color{black}}%
      \expandafter\def\csname LT1\endcsname{\color{black}}%
      \expandafter\def\csname LT2\endcsname{\color{black}}%
      \expandafter\def\csname LT3\endcsname{\color{black}}%
      \expandafter\def\csname LT4\endcsname{\color{black}}%
      \expandafter\def\csname LT5\endcsname{\color{black}}%
      \expandafter\def\csname LT6\endcsname{\color{black}}%
      \expandafter\def\csname LT7\endcsname{\color{black}}%
      \expandafter\def\csname LT8\endcsname{\color{black}}%
    \fi
  \fi
    \setlength{\unitlength}{0.0500bp}%
    \ifx\gptboxheight\undefined%
      \newlength{\gptboxheight}%
      \newlength{\gptboxwidth}%
      \newsavebox{\gptboxtext}%
    \fi%
    \setlength{\fboxrule}{0.5pt}%
    \setlength{\fboxsep}{1pt}%
\begin{picture}(10080.00,2040.00)%
    \gplgaddtomacro\gplbacktext{%
      \csname LTb\endcsname
      \put(849,595){\makebox(0,0)[r]{\strut{}\scriptsize $-80$}}%
      \csname LTb\endcsname
      \put(849,863){\makebox(0,0)[r]{\strut{}\scriptsize $-40$}}%
      \csname LTb\endcsname
      \put(849,1131){\makebox(0,0)[r]{\strut{}\scriptsize $0$}}%
      \csname LTb\endcsname
      \put(849,1399){\makebox(0,0)[r]{\strut{}\scriptsize $40$}}%
      \csname LTb\endcsname
      \put(849,1667){\makebox(0,0)[r]{\strut{}\scriptsize $80$}}%
      \csname LTb\endcsname
      \put(951,409){\makebox(0,0){\strut{}\scriptsize $0$}}%
      \csname LTb\endcsname
      \put(2421,409){\makebox(0,0){\strut{}\scriptsize $100$}}%
      \csname LTb\endcsname
      \put(3892,409){\makebox(0,0){\strut{}\scriptsize $200$}}%
      \csname LTb\endcsname
      \put(5362,409){\makebox(0,0){\strut{}\scriptsize $300$}}%
      \csname LTb\endcsname
      \put(6832,409){\makebox(0,0){\strut{}\scriptsize $400$}}%
      \csname LTb\endcsname
      \put(8303,409){\makebox(0,0){\strut{}\scriptsize $500$}}%
      \csname LTb\endcsname
      \put(9773,409){\makebox(0,0){\strut{}\scriptsize $600$}}%
    }%
    \gplgaddtomacro\gplfronttext{%
      \csname LTb\endcsname
      \put(459,1131){\rotatebox{-270}{\makebox(0,0){\strut{}Torque (\si{\newton\metre})}}}%
      \csname LTb\endcsname
      \put(5362,130){\makebox(0,0){\strut{}Time (s)}}%
      \csname LTb\endcsname
      \put(3619,1873){\makebox(0,0)[r]{\strut{}Combined + Feedback Torque}}%
      \csname LTb\endcsname
      \put(5733,1873){\makebox(0,0)[r]{\strut{}NP Torque}}%
      \csname LTb\endcsname
      \put(7847,1873){\makebox(0,0)[r]{\strut{}P Torque}}%
    }%
    \gplbacktext
    \put(0,0){\includegraphics{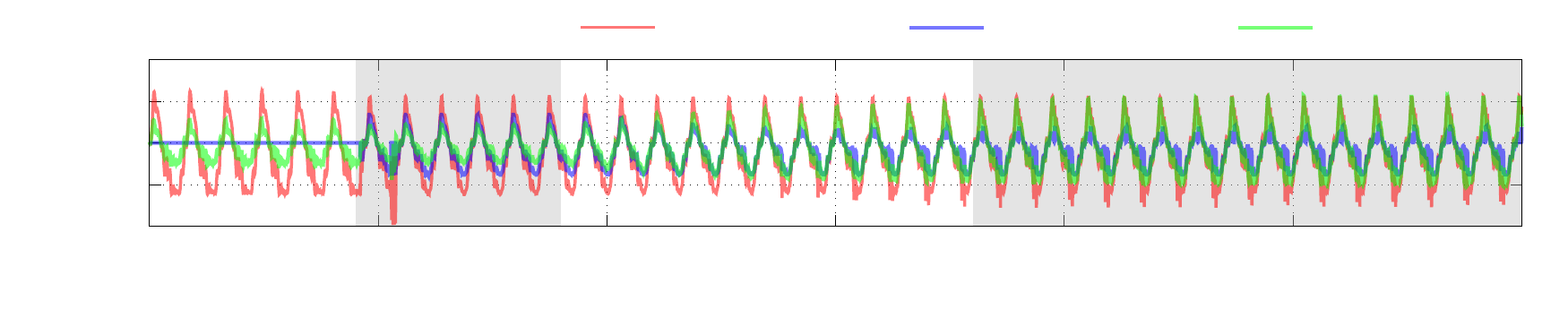}}%
    \gplfronttext
  \end{picture}%
\endgroup
\vspace*{1\baselineskip}

    \caption{Joint 2 - Torque Contributions - The contribution of both components to the combined output of the model with state feedback. With no consistency transform, we can see that the components stall in learning. The stall indicates that the change in the Parametric component is not taken into account in the Non-Parametric component. Phase changes indicated by background. Phase 1: No Parametric update, Non-Parametric updates but does not output. Phase 2: No Parametric update, Non-Parametric updates and outputs. Phase 3: Parametric updates, Non-Parametric updates but no transform is applied. Phase 4: Same as Phase 3.}\label{subfig:InconsistentTorque}
\end{subfigure}
\begin{subfigure}{\textwidth}
    \centering
\begingroup
  \makeatletter
  \providecommand\color[2][]{%
    \GenericError{(gnuplot) \space\space\space\@spaces}{%
      Package color not loaded in conjunction with
      terminal option `colourtext'%
    }{See the gnuplot documentation for explanation.%
    }{Either use 'blacktext' in gnuplot or load the package
      color.sty in LaTeX.}%
    \renewcommand\color[2][]{}%
  }%
  \providecommand\includegraphics[2][]{%
    \GenericError{(gnuplot) \space\space\space\@spaces}{%
      Package graphicx or graphics not loaded%
    }{See the gnuplot documentation for explanation.%
    }{The gnuplot epslatex terminal needs graphicx.sty or graphics.sty.}%
    \renewcommand\includegraphics[2][]{}%
  }%
  \providecommand\rotatebox[2]{#2}%
  \@ifundefined{ifGPcolor}{%
    \newif\ifGPcolor
    \GPcolortrue
  }{}%
  \@ifundefined{ifGPblacktext}{%
    \newif\ifGPblacktext
    \GPblacktexttrue
  }{}%
  \let\gplgaddtomacro\g@addto@macro
  \gdef\gplbacktext{}%
  \gdef\gplfronttext{}%
  \makeatother
  \ifGPblacktext
    \def\colorrgb#1{}%
    \def\colorgray#1{}%
  \else
    \ifGPcolor
      \def\colorrgb#1{\color[rgb]{#1}}%
      \def\colorgray#1{\color[gray]{#1}}%
      \expandafter\def\csname LTw\endcsname{\color{white}}%
      \expandafter\def\csname LTb\endcsname{\color{black}}%
      \expandafter\def\csname LTa\endcsname{\color{black}}%
      \expandafter\def\csname LT0\endcsname{\color[rgb]{1,0,0}}%
      \expandafter\def\csname LT1\endcsname{\color[rgb]{0,1,0}}%
      \expandafter\def\csname LT2\endcsname{\color[rgb]{0,0,1}}%
      \expandafter\def\csname LT3\endcsname{\color[rgb]{1,0,1}}%
      \expandafter\def\csname LT4\endcsname{\color[rgb]{0,1,1}}%
      \expandafter\def\csname LT5\endcsname{\color[rgb]{1,1,0}}%
      \expandafter\def\csname LT6\endcsname{\color[rgb]{0,0,0}}%
      \expandafter\def\csname LT7\endcsname{\color[rgb]{1,0.3,0}}%
      \expandafter\def\csname LT8\endcsname{\color[rgb]{0.5,0.5,0.5}}%
    \else
      \def\colorrgb#1{\color{black}}%
      \def\colorgray#1{\color[gray]{#1}}%
      \expandafter\def\csname LTw\endcsname{\color{white}}%
      \expandafter\def\csname LTb\endcsname{\color{black}}%
      \expandafter\def\csname LTa\endcsname{\color{black}}%
      \expandafter\def\csname LT0\endcsname{\color{black}}%
      \expandafter\def\csname LT1\endcsname{\color{black}}%
      \expandafter\def\csname LT2\endcsname{\color{black}}%
      \expandafter\def\csname LT3\endcsname{\color{black}}%
      \expandafter\def\csname LT4\endcsname{\color{black}}%
      \expandafter\def\csname LT5\endcsname{\color{black}}%
      \expandafter\def\csname LT6\endcsname{\color{black}}%
      \expandafter\def\csname LT7\endcsname{\color{black}}%
      \expandafter\def\csname LT8\endcsname{\color{black}}%
    \fi
  \fi
    \setlength{\unitlength}{0.0500bp}%
    \ifx\gptboxheight\undefined%
      \newlength{\gptboxheight}%
      \newlength{\gptboxwidth}%
      \newsavebox{\gptboxtext}%
    \fi%
    \setlength{\fboxrule}{0.5pt}%
    \setlength{\fboxsep}{1pt}%
\begin{picture}(10080.00,2040.00)%
    \gplgaddtomacro\gplbacktext{%
      \csname LTb\endcsname
      \put(951,595){\makebox(0,0)[r]{\strut{}\scriptsize $-1$}}%
      \csname LTb\endcsname
      \put(951,809){\makebox(0,0)[r]{\strut{}\scriptsize $-0.6$}}%
      \csname LTb\endcsname
      \put(951,1024){\makebox(0,0)[r]{\strut{}\scriptsize $-0.2$}}%
      \csname LTb\endcsname
      \put(951,1238){\makebox(0,0)[r]{\strut{}\scriptsize $0.2$}}%
      \csname LTb\endcsname
      \put(951,1453){\makebox(0,0)[r]{\strut{}\scriptsize $0.6$}}%
      \csname LTb\endcsname
      \put(951,1667){\makebox(0,0)[r]{\strut{}\scriptsize $1$}}%
      \csname LTb\endcsname
      \put(1053,409){\makebox(0,0){\strut{}\scriptsize $0$}}%
      \csname LTb\endcsname
      \put(2506,409){\makebox(0,0){\strut{}\scriptsize $100$}}%
      \csname LTb\endcsname
      \put(3960,409){\makebox(0,0){\strut{}\scriptsize $200$}}%
      \csname LTb\endcsname
      \put(5413,409){\makebox(0,0){\strut{}\scriptsize $300$}}%
      \csname LTb\endcsname
      \put(6866,409){\makebox(0,0){\strut{}\scriptsize $400$}}%
      \csname LTb\endcsname
      \put(8320,409){\makebox(0,0){\strut{}\scriptsize $500$}}%
      \csname LTb\endcsname
      \put(9773,409){\makebox(0,0){\strut{}\scriptsize $600$}}%
    }%
    \gplgaddtomacro\gplfronttext{%
      \csname LTb\endcsname
      \put(459,1131){\rotatebox{-270}{\makebox(0,0){\strut{}Error}}}%
      \csname LTb\endcsname
      \put(5413,130){\makebox(0,0){\strut{}Time (s)}}%
      \csname LTb\endcsname
      \put(4727,1873){\makebox(0,0)[r]{\strut{}Velocity Error ($\si{\radian\per\second}$)}}%
      \csname LTb\endcsname
      \put(7555,1873){\makebox(0,0)[r]{\strut{}Position Error ($\si{\radian}$)}}%
    }%
    \gplbacktext
    \put(0,0){\includegraphics{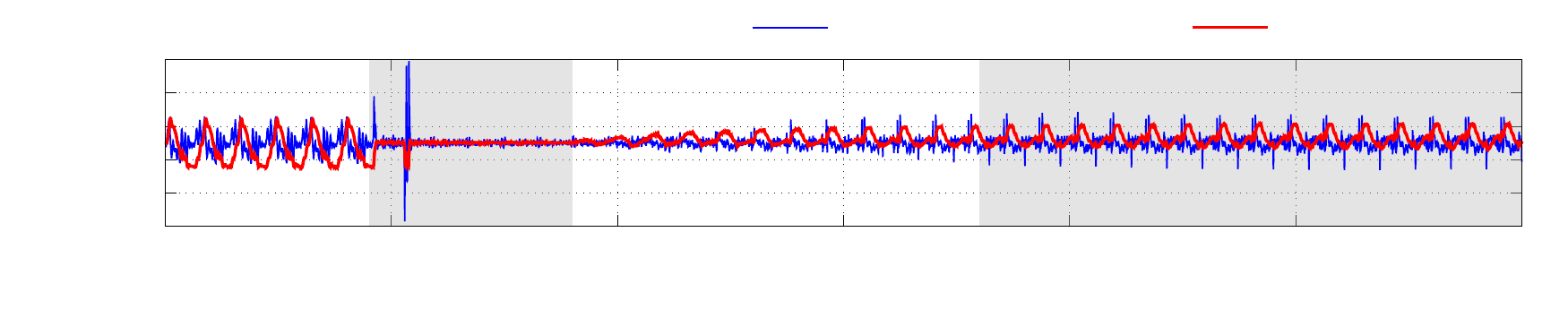}}%
    \gplfronttext
  \end{picture}%
\endgroup
\vspace*{1\baselineskip}

    \caption{Joint 2 - Position and Velocity Error - As the Parametric component continues to learn we see that the errors increase due to this inconsistency. Phase changes indicated by background.}
\end{subfigure}
\vspace*{-0.8\baselineskip}
\caption{Non-Consistent Transform Exemplar}\label{fig:InconsistentExemplar}
\vspace*{2\baselineskip}
    \begin{subfigure}{\textwidth}
    \centering
  
\begingroup
\makeatletter
\providecommand\color[2][]{%
  \GenericError{(gnuplot) \space\space\space\@spaces}{%
    Package color not loaded in conjunction with
    terminal option `colourtext'%
  }{See the gnuplot documentation for explanation.%
  }{Either use 'blacktext' in gnuplot or load the package
    color.sty in LaTeX.}%
  \renewcommand\color[2][]{}%
}%
\providecommand\includegraphics[2][]{%
  \GenericError{(gnuplot) \space\space\space\@spaces}{%
    Package graphicx or graphics not loaded%
  }{See the gnuplot documentation for explanation.%
  }{The gnuplot epslatex terminal needs graphicx.sty or graphics.sty.}%
  \renewcommand\includegraphics[2][]{}%
}%
\providecommand\rotatebox[2]{#2}%
\@ifundefined{ifGPcolor}{%
  \newif\ifGPcolor
  \GPcolortrue
}{}%
\@ifundefined{ifGPblacktext}{%
  \newif\ifGPblacktext
  \GPblacktexttrue
}{}%
\let\gplgaddtomacro\g@addto@macro
\gdef\gplbacktext{}%
\gdef\gplfronttext{}%
\makeatother
\ifGPblacktext
  \def\colorrgb#1{}%
  \def\colorgray#1{}%
\else
  \ifGPcolor
    \def\colorrgb#1{\color[rgb]{#1}}%
    \def\colorgray#1{\color[gray]{#1}}%
    \expandafter\def\csname LTw\endcsname{\color{white}}%
    \expandafter\def\csname LTb\endcsname{\color{black}}%
    \expandafter\def\csname LTa\endcsname{\color{black}}%
    \expandafter\def\csname LT0\endcsname{\color[rgb]{1,0,0}}%
    \expandafter\def\csname LT1\endcsname{\color[rgb]{0,1,0}}%
    \expandafter\def\csname LT2\endcsname{\color[rgb]{0,0,1}}%
    \expandafter\def\csname LT3\endcsname{\color[rgb]{1,0,1}}%
    \expandafter\def\csname LT4\endcsname{\color[rgb]{0,1,1}}%
    \expandafter\def\csname LT5\endcsname{\color[rgb]{1,1,0}}%
    \expandafter\def\csname LT6\endcsname{\color[rgb]{0,0,0}}%
    \expandafter\def\csname LT7\endcsname{\color[rgb]{1,0.3,0}}%
    \expandafter\def\csname LT8\endcsname{\color[rgb]{0.5,0.5,0.5}}%
  \else
    \def\colorrgb#1{\color{black}}%
    \def\colorgray#1{\color[gray]{#1}}%
    \expandafter\def\csname LTw\endcsname{\color{white}}%
    \expandafter\def\csname LTb\endcsname{\color{black}}%
    \expandafter\def\csname LTa\endcsname{\color{black}}%
    \expandafter\def\csname LT0\endcsname{\color{black}}%
    \expandafter\def\csname LT1\endcsname{\color{black}}%
    \expandafter\def\csname LT2\endcsname{\color{black}}%
    \expandafter\def\csname LT3\endcsname{\color{black}}%
    \expandafter\def\csname LT4\endcsname{\color{black}}%
    \expandafter\def\csname LT5\endcsname{\color{black}}%
    \expandafter\def\csname LT6\endcsname{\color{black}}%
    \expandafter\def\csname LT7\endcsname{\color{black}}%
    \expandafter\def\csname LT8\endcsname{\color{black}}%
  \fi
\fi
  \setlength{\unitlength}{0.0500bp}%
  \ifx\gptboxheight\undefined%
    \newlength{\gptboxheight}%
    \newlength{\gptboxwidth}%
    \newsavebox{\gptboxtext}%
  \fi%
  \setlength{\fboxrule}{0.5pt}%
  \setlength{\fboxsep}{1pt}%
\begin{picture}(10080.00,2040.00)%
  \gplgaddtomacro\gplbacktext{%
    \csname LTb\endcsname
    \put(849,595){\makebox(0,0)[r]{\strut{}\scriptsize $-80$}}%
    \csname LTb\endcsname
    \put(849,863){\makebox(0,0)[r]{\strut{}\scriptsize $-40$}}%
    \csname LTb\endcsname
    \put(849,1131){\makebox(0,0)[r]{\strut{}\scriptsize $0$}}%
    \csname LTb\endcsname
    \put(849,1399){\makebox(0,0)[r]{\strut{}\scriptsize $40$}}%
    \csname LTb\endcsname
    \put(849,1667){\makebox(0,0)[r]{\strut{}\scriptsize $80$}}%
    \csname LTb\endcsname
    \put(951,409){\makebox(0,0){\strut{}\scriptsize $0$}}%
    \csname LTb\endcsname
    \put(2421,409){\makebox(0,0){\strut{}\scriptsize $100$}}%
    \csname LTb\endcsname
    \put(3892,409){\makebox(0,0){\strut{}\scriptsize $200$}}%
    \csname LTb\endcsname
    \put(5362,409){\makebox(0,0){\strut{}\scriptsize $300$}}%
    \csname LTb\endcsname
    \put(6832,409){\makebox(0,0){\strut{}\scriptsize $400$}}%
    \csname LTb\endcsname
    \put(8303,409){\makebox(0,0){\strut{}\scriptsize $500$}}%
    \csname LTb\endcsname
    \put(9773,409){\makebox(0,0){\strut{}\scriptsize $600$}}%
  }%
  \gplgaddtomacro\gplfronttext{%
    \csname LTb\endcsname
    \put(459,1131){\rotatebox{-270}{\makebox(0,0){\strut{}Torque (\si{\newton\metre})}}}%
    \csname LTb\endcsname
    \put(5362,130){\makebox(0,0){\strut{}Time (s)}}%
    \csname LTb\endcsname
    \put(3619,1873){\makebox(0,0)[r]{\strut{}Combined + Feedback Torque}}%
    \csname LTb\endcsname
    \put(5733,1873){\makebox(0,0)[r]{\strut{}NP Torque}}%
    \csname LTb\endcsname
    \put(7847,1873){\makebox(0,0)[r]{\strut{}P Torque}}%
  }%
  \gplbacktext
  \put(0,0){\includegraphics{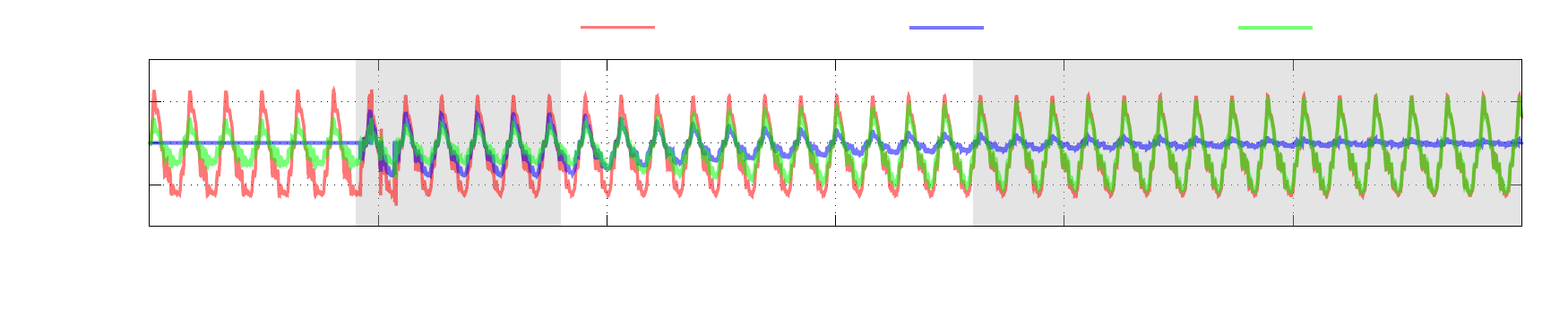}}%
  \gplfronttext
\end{picture}%
\endgroup
\vspace*{1\baselineskip}
    \caption{Joint 2 - Torque Contributions - The contribution of both components to the combined output with state feedback. We see that as the components adapt, the Parametric component increases and the Non-Parametric component equivalently decreases. Phase changes indicated by background. Phase 1: No Parametric update, Non-Parametric updates but does not output. Phase 2: No Parametric update, Non-Parametric updates and outputs. Phase 3: Parametric updates, Non-Parametric does not update but the consistency transform is applied. Phase 4: Parametric updates, Non-Parametric updates and the consistency transform are applied.}\label{subfig:ConsistentTorque}
\end{subfigure}
\begin{subfigure}{\textwidth}
    \centering
\begingroup
  \makeatletter
  \providecommand\color[2][]{%
    \GenericError{(gnuplot) \space\space\space\@spaces}{%
      Package color not loaded in conjunction with
      terminal option `colourtext'%
    }{See the gnuplot documentation for explanation.%
    }{Either use 'blacktext' in gnuplot or load the package
      color.sty in LaTeX.}%
    \renewcommand\color[2][]{}%
  }%
  \providecommand\includegraphics[2][]{%
    \GenericError{(gnuplot) \space\space\space\@spaces}{%
      Package graphicx or graphics not loaded%
    }{See the gnuplot documentation for explanation.%
    }{The gnuplot epslatex terminal needs graphicx.sty or graphics.sty.}%
    \renewcommand\includegraphics[2][]{}%
  }%
  \providecommand\rotatebox[2]{#2}%
  \@ifundefined{ifGPcolor}{%
    \newif\ifGPcolor
    \GPcolortrue
  }{}%
  \@ifundefined{ifGPblacktext}{%
    \newif\ifGPblacktext
    \GPblacktexttrue
  }{}%
  \let\gplgaddtomacro\g@addto@macro
  \gdef\gplbacktext{}%
  \gdef\gplfronttext{}%
  \makeatother
  \ifGPblacktext
    \def\colorrgb#1{}%
    \def\colorgray#1{}%
  \else
    \ifGPcolor
      \def\colorrgb#1{\color[rgb]{#1}}%
      \def\colorgray#1{\color[gray]{#1}}%
      \expandafter\def\csname LTw\endcsname{\color{white}}%
      \expandafter\def\csname LTb\endcsname{\color{black}}%
      \expandafter\def\csname LTa\endcsname{\color{black}}%
      \expandafter\def\csname LT0\endcsname{\color[rgb]{1,0,0}}%
      \expandafter\def\csname LT1\endcsname{\color[rgb]{0,1,0}}%
      \expandafter\def\csname LT2\endcsname{\color[rgb]{0,0,1}}%
      \expandafter\def\csname LT3\endcsname{\color[rgb]{1,0,1}}%
      \expandafter\def\csname LT4\endcsname{\color[rgb]{0,1,1}}%
      \expandafter\def\csname LT5\endcsname{\color[rgb]{1,1,0}}%
      \expandafter\def\csname LT6\endcsname{\color[rgb]{0,0,0}}%
      \expandafter\def\csname LT7\endcsname{\color[rgb]{1,0.3,0}}%
      \expandafter\def\csname LT8\endcsname{\color[rgb]{0.5,0.5,0.5}}%
    \else
      \def\colorrgb#1{\color{black}}%
      \def\colorgray#1{\color[gray]{#1}}%
      \expandafter\def\csname LTw\endcsname{\color{white}}%
      \expandafter\def\csname LTb\endcsname{\color{black}}%
      \expandafter\def\csname LTa\endcsname{\color{black}}%
      \expandafter\def\csname LT0\endcsname{\color{black}}%
      \expandafter\def\csname LT1\endcsname{\color{black}}%
      \expandafter\def\csname LT2\endcsname{\color{black}}%
      \expandafter\def\csname LT3\endcsname{\color{black}}%
      \expandafter\def\csname LT4\endcsname{\color{black}}%
      \expandafter\def\csname LT5\endcsname{\color{black}}%
      \expandafter\def\csname LT6\endcsname{\color{black}}%
      \expandafter\def\csname LT7\endcsname{\color{black}}%
      \expandafter\def\csname LT8\endcsname{\color{black}}%
    \fi
  \fi
    \setlength{\unitlength}{0.0500bp}%
    \ifx\gptboxheight\undefined%
      \newlength{\gptboxheight}%
      \newlength{\gptboxwidth}%
      \newsavebox{\gptboxtext}%
    \fi%
    \setlength{\fboxrule}{0.5pt}%
    \setlength{\fboxsep}{1pt}%
\begin{picture}(10080.00,2040.00)%
    \gplgaddtomacro\gplbacktext{%
      \csname LTb\endcsname
      \put(951,595){\makebox(0,0)[r]{\strut{}\scriptsize $-1$}}%
      \csname LTb\endcsname
      \put(951,809){\makebox(0,0)[r]{\strut{}\scriptsize $-0.6$}}%
      \csname LTb\endcsname
      \put(951,1024){\makebox(0,0)[r]{\strut{}\scriptsize $-0.2$}}%
      \csname LTb\endcsname
      \put(951,1238){\makebox(0,0)[r]{\strut{}\scriptsize $0.2$}}%
      \csname LTb\endcsname
      \put(951,1453){\makebox(0,0)[r]{\strut{}\scriptsize $0.6$}}%
      \csname LTb\endcsname
      \put(951,1667){\makebox(0,0)[r]{\strut{}\scriptsize $1$}}%
      \csname LTb\endcsname
      \put(1053,409){\makebox(0,0){\strut{}\scriptsize $0$}}%
      \csname LTb\endcsname
      \put(2506,409){\makebox(0,0){\strut{}\scriptsize $100$}}%
      \csname LTb\endcsname
      \put(3960,409){\makebox(0,0){\strut{}\scriptsize $200$}}%
      \csname LTb\endcsname
      \put(5413,409){\makebox(0,0){\strut{}\scriptsize $300$}}%
      \csname LTb\endcsname
      \put(6866,409){\makebox(0,0){\strut{}\scriptsize $400$}}%
      \csname LTb\endcsname
      \put(8320,409){\makebox(0,0){\strut{}\scriptsize $500$}}%
      \csname LTb\endcsname
      \put(9773,409){\makebox(0,0){\strut{}\scriptsize $600$}}%
    }%
    \gplgaddtomacro\gplfronttext{%
      \csname LTb\endcsname
      \put(459,1131){\rotatebox{-270}{\makebox(0,0){\strut{}Error}}}%
      \csname LTb\endcsname
      \put(5413,130){\makebox(0,0){\strut{}Time (s)}}%
      \csname LTb\endcsname
      \put(4727,1873){\makebox(0,0)[r]{\strut{}Velocity Error ($\si{\radian\per\second}$)}}%
      \csname LTb\endcsname
      \put(7555,1873){\makebox(0,0)[r]{\strut{}Position Error ($\si{\radian}$)}}%
    }%
    \gplbacktext
    \put(0,0){\includegraphics{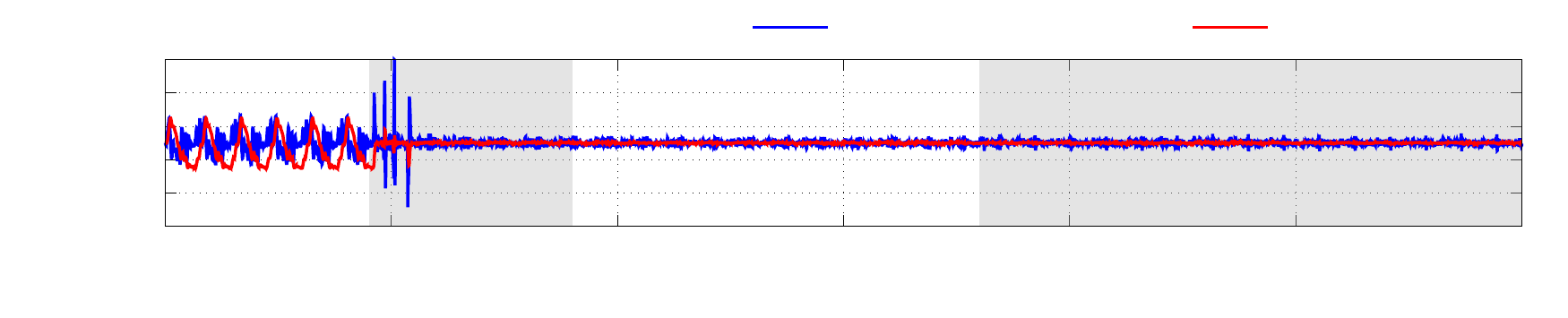}}%
    \gplfronttext
  \end{picture}%
\endgroup
\vspace*{1\baselineskip}
    \caption{Joint 2 - Position and Velocity Error - We determine that the errors during the learning do not change significantly when consistently transforming the model. Phase changes indicated by background.}
\end{subfigure}
\caption{Consistent Transform Exemplar}\label{fig:ConsistentExemplar}
\end{figure*}

The error behaviour is more evident with the trajectory shown in \cref{fig:snaps}. The trajectory provides a specific run's evolution over the experiment, shown in \cref{fig:InconsistentExemplar,fig:ConsistentExemplar}. \Cref{fig:InconsistentExemplar}, where the transform is not applied, shows that the learning appears to stall, with the sum of the two components being much greater than the output of the controller indicating high feedback torques. The deteriorating trajectory tracking also indicates the higher feedback torques needed to compensate for the incorrect model torque.

\Cref{fig:ConsistentExemplar}, where the transform is applied, maintains the trajectory tracking performance whilst learning both components. \Cref{subfig:ConsistentTorque} shows clearly the transference behaviour between the two components. We can see that as the Parametric component gets updated iteratively, the torque contribution is increased. With the Non-Parametric component we can see a relative decrease in torque contribution of a similar magnitude as the increase in Parametric torque. This behaviour indicates that the components are being adapted with respect to each other so that their output is consistent, and the training biases the model towards the Parametric component. 

An observation from the experiments is that the bias has the effect of simplifying the Non-Parametric target function, as evidenced by the reduced increase of sub-models in the GMM. Without the transform, we observed the GMM increased by eight sub-models on average, whereas with the transform the GMM increased by a single sub-model on average. This indicates that the GMM is kept consistent as fewer new sub-models are needed for the new target function.

\section{Conclusion}
\label{sec:conclusion}
\vspace*{-1mm}
This paper addresses the non-stationary issue when updating both components of a Semi-Parametric model simultaneously and online. We have verified that a consistency transform of the Non-Parametric component with respect to changes in parameters can reduce errors caused by the non-stationary problem.

Overall, from \cref{ssec:RPV,ssec:TV,fig:ConsistentExemplar,fig:InconsistentExemplar}, we show using the consistency transform we are able to maintain performance whilst adapting both components online without explicit retraining of the Non-Parametric component. The experiments also show that the Semi-Parametric model biases the learning towards the Parametric component, hence reducing the contribution of the Non-Parametric component. The bias in the Semi-Parametric model should allow greater ability in terms of generalization, as the Parametric component typically does a better job at generalizing over the state space. The reduced contribution of the Non-Parametric component will also reduce the effect of incorrect predictions as they should be relatively small compared to the Parametric component.

\section{Discussion}\balance
In future we plan on expanding the analysis of the simplification of the Non-Parametric target function and the generalization capability of the model due to the bias effect towards the Parametric component. Which has been observed during the experiments in \cref{ssec:RPV,ssec:TV}.

The issues around the Gaussians, becoming singular or becoming numerically unstable, can be further explored as an extension. Possible solutions to explore are; adding robustness to singularities, or regularization, to the GMM sub-models, adapting the idea of consistency transform to other models, such as neural networks that are better suited for pure regression problems. Another extension would be looking into the physical plausibility of the inertial parameters, as this is currently not enforced.


\bibliography{IEEEabrv,references}


\end{document}